\documentclass[onecolumn]{article}
\PassOptionsToPackage{reqno}{amsmath}
\usepackage[letterpaper]{geometry}

\usepackage{siamproceedings}

\usepackage[T1]{fontenc}
\usepackage{amsfonts}
\usepackage{graphicx}
\usepackage{epstopdf}
\usepackage{enumitem}
\usepackage{algorithmic}
\usepackage{microtype}
\usepackage{subfigure}
\usepackage{booktabs}
\usepackage{amsmath}
\usepackage{amssymb}
\usepackage{mathtools}
\usepackage{placeins}
\usepackage{comment}
\usepackage{float}
\ifpdf
  \DeclareGraphicsExtensions{.eps,.pdf,.png,.jpg}
\else
  \DeclareGraphicsExtensions{.eps}
\fi

\newsiamremark{remark}{Remark}
\newsiamremark{hypothesis}{Hypothesis}
\crefname{hypothesis}{Hypothesis}{Hypotheses}
\newsiamthm{claim}{Claim}

\usepackage{amsopn}

\begin{document}

\newcommand\relatedversion{}

\title{\Large Multilevel Graph Wavelet Compressed Sensing with Scale-Aware Neural Recovery\relatedversion}
\author{Amirhossein Nouranizadeh$^{+}$\thanks{Simplicial Technologies Inc., Toronto, ON, Canada (\email{\mbox{amirhossein@simplicial.ai}}).}
\and Sarang Patil$^{+}$\thanks{Department of Data Science, New Jersey Institute of Technology, Newark, NJ (\email{sp3463@njit.edu}).}
\and Alan John Varghese\thanks{School of Engineering, Brown University, Providence, RI (\email{alan\_john\_varghese@brown.edu}).}
\and Varsha Narayanan\thanks{Department of Computer Science, New Jersey Institute of Technology, Newark, NJ (\email{vn299@njit.edu}).}
\and Amit  Chakraborty\thanks{Siemens Foundational Technology, Princeton, NJ (\email{amit.chakraborty@siemens.com}).}
\and Mengjia Xu\thanks{Department of Data Science, New Jersey Institute of Technology, Newark, NJ (corresponding author: \email{mx6@njit.edu}).}}

\date{}

\maketitle
\let\thefootnote\relax\footnotetext{$^{+}$\, A. Nouranizadeh and S. Patil contributed equally to this work.}





\begin{abstract}
Scientific machine learning methods such as neural operators and physics-informed neural networks have advanced engineering applications and inverse problems, but their training typically requires large volumes of simulated data.
This makes data preparation and model training expensive.
We propose Graph Wavelet Compressed Sensing (GWCS), a learning-based framework for offline compression of graph signals by representing them as sparse, interpretable wavelet-domain representations using the spectral graph wavelet transform.
The framework combines a nonparametric multilevel importance sampler, which retains high-energy wavelet coefficients within each scale for a given compression ratio, with a scale-aware graph neural network that reconstructs the signal from the sparse coefficients.
We evaluate the proposed framework on synthetic approximately band-limited graph signals over random graphs and four PDE simulation datasets over meshes, which include Turbulent Radiative Layer, Viscoelastic Instability, Kolmogorov Flow, and Dynamic Stall.
We compare against graph signal sampling methods and graph autoencoder baselines.
Results demonstrate that the framework achieves high reconstruction fidelity and substantial data compression compared to existing benchmarks.~\footnote{Code available at \url{https://github.com/amrhssn/gwcs/}.}
\end{abstract}

\section{Introduction}
\label{introduction}
Scientific machine learning (SciML) has emerged as a powerful paradigm for modeling complex physical, biological, and engineering systems. By integrating physics-based principles with data-driven inference, methods such as neural operators~\cite{lu2021learning,kovachki2023neural} and physics-informed neural networks (PINNs)~\cite{raissi2019physics} have enabled the efficient approximation of partial differential equations (PDEs) and the solution of inverse problems across diverse scientific domains. However, these approaches typically rely on large volumes of high-fidelity simulation data for training, particularly when learning operators on fine spatial and temporal resolutions. This leads to substantial computational and storage costs, making scalability a major challenge for high-dimensional spatiotemporal systems. To address this, there has been a growing interest in learning efficient low-dimensional latent representations that can capture the essential structure of the underlying dynamics. This becomes even more challenging in settings involving unstructured data, such as meshes or irregular domains, which are naturally represented as graphs. Compressed sensing (CS), a well-established mathematical theory widely used in domains such as medical image acquisition, exploits the underlying structure of natural signals to reconstruct an entire
image from a few transform-domain samples by solving an optimization
problem \cite{candes2006robust, donoho2006compressed}.
In the graph signal processing domain, analogous graph signal sampling (GSS) techniques reconstruct graph signals from a few node- or spectral-domain
samples by solving an optimization problem, typically assuming
prior knowledge of the spectral properties of the signal under
study \cite{ortega2022introduction}.
These techniques motivate the design of a compression
scheme that exploits the transform-domain sparsity already known
to hold for physical fields on meshes and graphs
\cite{schneider2010wavelet, ricaud2013sparsity}, without requiring
per-signal optimization or prior knowledge of the signal's
bandwidth.

In this work, inspired by CS and GSS theories, we propose Graph Wavelet Compressed Sensing (GWCS),
a learning-based, offline graph signal compression technique that operates
in the wavelet domain of graph-structured data.
GWCS exploits the multiscale sparsity of graph wavelet coefficients to compress a signal through
multilevel importance sampling of the coefficients,
and trains a graph neural network (GNN) to reconstruct the full signal from the resulting sparse coefficients.

\textbf{The main contributions are as follows.}
\begin{enumerate}[leftmargin=*, label=(\roman*), topsep=1pt, itemsep=0pt]
  \item We propose \textbf{Graph Wavelet Compressed Sensing (GWCS)}, a
  framework for learning-based compressed sensing of graph signals that combines
  a nonparametric \textbf{Multilevel Importance Sampling (MLIS)} module, which retains
  high-energy wavelet coefficients across scales, with a neural recovery
  model.
  \item Within GWCS, we introduce the \textbf{Neural Inverse Graph Wavelet
  Transform (NIGWT)}, a scale-aware Encode-Process-Decode GNN that uses learnable scale
  embeddings and a residual IGWT connection to
  reconstruct signals from the sparse coefficients retained by MLIS.
  \item To isolate the contribution of the wavelet domain, we also
  develop the \textbf{Sparse Graph Autoencoder (SGAE)}, a wavelet-free method with a
  learned sparse bottleneck, which we use throughout the experiments as a
  strong baseline for comparison against GWCS.
\end{enumerate}

\section{Related Works}
\label{sec:related-works}
\subsection{Compressed Sensing.} Compressed sensing (CS) has emerged as a fundamental framework for recovering signals from sub-Nyquist measurements by exploiting low-dimensional structure such as sparsity~\cite{donoho2006compressed, candes2006robust}. Under suitable conditions, namely sparsity in a transform domain and incoherence between sampling and representation bases, accurate reconstruction can be achieved from a number of measurements proportional to the intrinsic sparsity of the signal, typically via $\ell_1$-regularized optimization.
These ideas have enabled a wide range of applications, including MRI, computerized tomography, and electron microscopy \cite{lustig2007sparse, chen2008prior, leary2013compressed}.
Real-world signals are only approximately sparse, incoherence conditions are rarely satisfied, and optimal sampling strategies are generally structured or variable-density rather than uniform random. These observations have motivated the community to extend the traditional CS theory into a new CS theory based on generalized assumptions of asymptotic sparsity, asymptotic incoherence, and multilevel random sampling \cite{adcock2017breaking, adcock2015quest}. These works highlight the importance of structured representations, particularly wavelets, and adaptive sampling strategies.

Our goal is to compress graph signals, drawing inspiration from the multilevel sampling framework of CS theory. Unlike typical CS settings, where sampling occurs online at acquisition time and reconstruction is performed by solving a per-sample $\ell_1$-regularized optimization problem, GWCS assumes access to the entire signal and its transform coefficients offline, and uses multilevel importance sampling to retain coefficients at multiple scales of the graph wavelet transform, and reconstructs the signal with a neural architecture. This enables compression for storage, transmission, and downstream processing.

\subsection{Graph Signal Sampling.} Graph signal sampling (GSS) is a family of graph signal processing techniques developed for sampling and reconstructing signals defined on irregular domains, where the goal is to select a subset of nodes that enable proper reconstruction of the full signal.
Most approaches assume that graph signals are smooth or approximately band-limited with respect to a graph operator, and design sampling strategies in either the spectral or vertex domain~\cite{chen2015discrete, anis2016efficient, marques2015sampling}.
Representative methods include greedy selection schemes, spectral proxy-based approaches that avoid eigendecomposition~\cite{anis2016efficient}, and randomized strategies such as determinantal point processes~\cite{tremblay2017graph}, often accompanied by theoretical guarantees on recovery error. Despite their effectiveness, these methods exhibit several limitations.
First, performance is highly sensitive to hyperparameters such as the assumed signal bandwidth or cutoff frequency, which are rarely known \emph{a priori}.
Second, they rely on fixed signal models (e.g., smoothness or band-limitedness), limiting their ability to adapt to complex, domain-specific structures.
Finally, many approaches require per-signal optimization or expensive preprocessing, which hinders scalability to large datasets.



\subsection{Graph Neural Networks.}
Graph neural networks (GNNs) provide a flexible framework for learning representations of graph-structured data through message passing~\cite{kipf2017semi, hamilton2017inductive}.
In the context of graph signals, GNNs have been used for reconstruction, denoising, and spatiotemporal prediction by learning data-driven priors that go beyond classical smoothness assumptions.
Related directions include graph autoencoders for learning compact latent node representations~\cite{kipf2016variational}, unrolled optimization networks for interpretable reconstruction~\cite{chen2021graph}, and multi-scale architectures that capture hierarchical dependencies~\cite{ying2018hierarchical, gao2019graph, nouranizadeh2021maximum}.
In many SciML applications, the underlying data naturally resides on unstructured meshes that can be represented as graphs, making graph-based compression and reconstruction techniques particularly relevant.
Our work builds on these ideas by combining multiscale graph representations with learning-based compressed sensing to enable efficient storage and reconstruction of scientific signals.

\section{Preliminaries}
\label{preliminaries}

\subsection{Graph Spectral Foundations.}
We consider a data set of $M$ samples, each consisting of an undirected graph and an associated signal, i.e.,
$\mathcal{D}=\{(\mathcal{G}^{(1)}, \mathbf{x}^{(1)}), \dots, (\mathcal{G}^{(M)}, \mathbf{x}^{(M)})\}$.
Each graph is defined as $\mathcal{G}^{(j)} = (\mathcal{V}^{(j)}, \mathcal{E}^{(j)})$,
where $\mathcal{V}^{(j)}$ and $\mathcal{E}^{(j)}$ denote the node and edge sets; $N^{(j)} = |\mathcal{V}^{(j)}|$ is the number of nodes. Each graph is associated with a graph signal $\mathbf{x}^{(j)} \in \mathbb{R}^{N^{(j)}}$ defined over its nodes, where each entry corresponds to the feature value at a node.
To simplify notation, we omit the sample index $j$ and make it explicit when needed.

The symmetric normalized graph Laplacian matrix is defined as
$\mathbf{L}_{\mathrm{norm}}=\mathbf{I}-\mathbf{D}^{-1/2} \mathbf{A} \mathbf{D}^{-1/2} \in \mathbb{R}^{N \times N}$, where $\mathbf{I}$ is the identity matrix, $\mathbf{A}$ is the adjacency matrix, and
$\mathbf{D}$ is the diagonal degree matrix.
The symmetric normalized Laplacian matrix has spectral decomposition:
$\mathbf{L}_{\mathrm{norm}}=\mathbf{U} \mathbf{\Lambda} \mathbf{U}^\top$,
where $\mathbf{U}=\left[ \mathbf{u}_1 \dots \mathbf{u}_N \right] \in \mathbb{R}^{N \times N}$ is the orthonormal matrix of eigenvectors and
$\mathbf{\Lambda}=\mathrm{diag}(\lambda_1, \dots, \lambda_N) \in \mathbb{R}^{N \times N}$ is the diagonal matrix of eigenvalues and $0 = \lambda_1 \leq \dots \leq \lambda_N \leq 2$. The eigenvectors form the graph Fourier basis, and the eigenvalues represent frequencies: small values correspond to smooth (low-frequency) signals over the graph, while large values correspond to rapidly varying (high-frequency) signals. The graph Fourier transform projects a graph signal $\mathbf{x}$ onto the spectral basis as $\tilde{\mathbf{x}} = \mathbf{U}^\top \mathbf{x}$, while the inverse reconstructs the signal via $\mathbf{x} = \mathbf{U}\tilde{\mathbf{x}}$~\cite{shuman2013emerging}. Graph spectral filtering is performed by applying a spectral filter $g: \mathbb{R} \rightarrow \mathbb{R}$ to the transformed signal, yielding $\tilde{\mathbf{y}} = g(\mathbf{\Lambda}) \tilde{\mathbf{x}}$, where $g(\mathbf{\Lambda}) = \mathrm{diag}(g(\lambda_1), \dots, g(\lambda_N))$. The filtered signal is then mapped back to the vertex domain as $\mathbf{y} = \mathbf{U}\tilde{\mathbf{y}} = \mathbf{U} g(\mathbf{\Lambda}) \mathbf{U}^\top \mathbf{x}$.

\subsection{Spectral Graph Wavelet Transform (SGWT).}
As a generalization of classical wavelet analysis to signals supported on irregular, non-Euclidean domains, the SGWT replaces geometric translation and scaling with filtering operations in the spectral domain of a graph operator such as the normalized Laplacian \cite{hammond2011wavelets}.
This enables a localized, multiscale analysis of graph signals without relying on an underlying Euclidean structure. In particular, SGWT constructs a spectral filterbank that decomposes signals into low-frequency components capturing global structure and high-frequency components capturing localized variations. 
To this end, we define a low-pass kernel $h: \mathbb{R} \rightarrow \mathbb{R}$ and a base band-pass kernel $g: \mathbb{R} \rightarrow \mathbb{R}$. From $g$ we construct $J$ band-pass kernels $g_{s}: \mathbb{R} \rightarrow \mathbb{R}$, $g_{s}(\lambda):=g(s\lambda)$, each parameterized by a scale value $s \in \{s_1, \ldots, s_J\}$, where $J$ is the number of band-pass scales. Each kernel satisfies $g_{s}(0)=0$, $\lim_{\lambda \rightarrow \infty}g_{s}(\lambda)=0$, and meets the admissibility condition of Lemma~5.1 in~\cite{hammond2011wavelets}.
Given the spectral decomposition of the normalized Laplacian, the graph wavelet filterbank consists of a low-pass filter $\mathbf{\Psi}_0 = \mathbf{U}h(\mathbf{\Lambda})\mathbf{U}^\top$, and
$J$ band-pass filters $\mathbf{\Psi}_{s_1}, \ldots, \mathbf{\Psi}_{s_J}$ where $\mathbf{\Psi}_{s_n} = \mathbf{U}g_{s_n}(\mathbf{\Lambda})\mathbf{U}^\top$. Stacking these graph filters yields the forward SGWT operator $\mathbf{\Psi}=\big[ \mathbf{\Psi}_0^\top, \mathbf{\Psi}_{s_1}^\top, \dots, \mathbf{\Psi}_{s_J}^\top\big]^\top \in \mathbb{R}^{FN \times N}$, where $F=1+J$ is the number of filters in the wavelet filterbank.
The graph wavelet coefficients are then obtained as $\mathbf{w}=\mathbf{\Psi}\mathbf{x} \in \mathbb{R}^{FN}$, which can be reshaped as a 2D matrix $\mathbf{W} \in \mathbb{R}^{N \times F}$, where each node has $F$ graph wavelet coefficients across different frequency bands. The graph signal can be recovered by the inverse SGWT, i.e., $\mathbf{x}=\mathbf{\Psi}^\dagger \mathbf{w}$, where $\mathbf{\Psi}^\dagger:=(\mathbf{\Psi}^\top\mathbf{\Psi})^{-1}\mathbf{\Psi}^\top$ is the pseudo-inverse of $\mathbf{\Psi}$.

\subsection{Chebyshev polynomial approximation of graph wavelet coefficients.}
Spectral filtering in SGWT requires the full eigendecomposition of the normalized graph Laplacian with cubic computational cost, which makes it impractical for large graphs.
In practice, we use truncated Chebyshev polynomial series to approximate spectral filters $h(\lambda)$ and $g_{s_j}(\lambda)$ as polynomial filters of normalized Laplacian that can be applied directly to the graph signal in the node domain efficiently \cite{hammond2011wavelets}, see the Appendix \ref{chebyshev} for computational details.

\section{Method}
\label{method}
\begin{figure*}[h]
  \centering
  \includegraphics[width=0.90\textwidth]{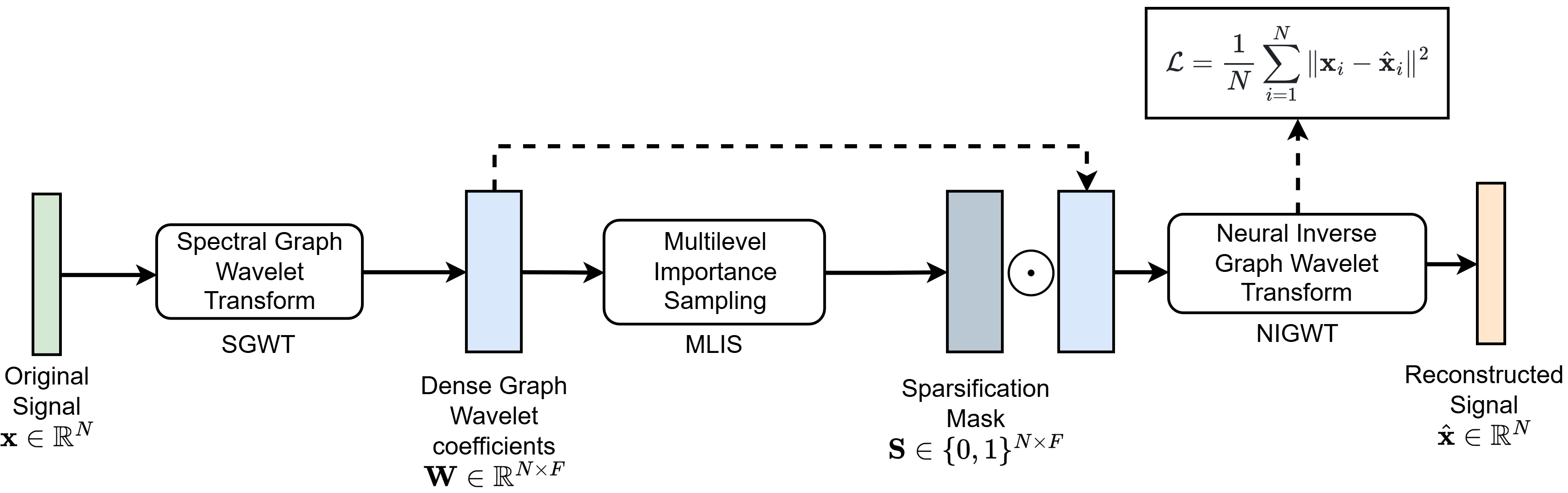}
  \vspace{-.18in}
  \caption{Overview of the GWCS Framework.
  1) The input graph signal $\mathbf{x}\in\mathbb{R}^{N}$ is first transformed by the Spectral Graph Wavelet Transform (SGWT), implemented via a Chebyshev approximation of degree $K=100$, to obtain a dense wavelet coefficient matrix $\mathbf{W}\in\mathbb{R}^{N\times F}$ ($F=5$ corresponds to $1$ lowpass and $J=4$ band-pass filters).
  2) The Multilevel Importance Sampling (MLIS) module allocates a retention
  budget of $m=cN$ coefficients across scales in proportion to each
  scale's energy (Eq.~\eqref{eq:budget_alloc}), then retains coefficients
  within each scale via importance sampling with probability proportional to
  squared coefficient magnitude (Eq.~\eqref{eq:sampling_dist}), yielding the
  binary sparsification mask $\mathbf{S}\in\{0,1\}^{N\times F}$ and the
  sparse coefficient matrix $\tilde{\mathbf{W}} \in \mathbb{R}^{N \times F}$.
  3) The Neural Inverse Graph Wavelet Transform (NIGWT) module recovers the full signal $\hat{\mathbf{x}}\in\mathbb{R}^{N}$ via an ``Encode-Process-Decode'' architecture with a residual IGWT connection.}
  \label{fig:nigwt_architecture}
\end{figure*}

In this section, we provide a detailed description of the proposed Graph Wavelet Compressed Sensing (GWCS) framework and Sparse Graph Autoencoder (SGAE). GWCS and SGAE are both controlled by a single \textbf{target compression ratio} $c \in (0, 1]$, which is the fraction of wavelet coefficients (GWCS) or latent entries (SGAE) kept relative to the size of the input signal.
For a graph signal with $N$ nodes and $C$ channels, this gives a retention budget $m=cNC$. We set $c=5\%$ for all experiments conducted in this paper. The GWCS framework in Figure~\ref{fig:nigwt_architecture} includes two main modules: (i) a nonparametric {\it Multilevel Importance Sampling (MLIS)} module that retains a fixed budget of wavelet coefficients using an energy-based heuristic informed by the multilevel structure of the wavelet transform, and, (ii) the {\it Neural Inverse Graph Wavelet Transform (NIGWT)} module, a scale-aware GNN recovery model that reconstructs the full signal from the sparse coefficients.

\subsection{Multilevel Importance Sampling Module.}
Inspired by the multilevel random subsampling technique introduced in the compressed sensing theory~\cite{adcock2017breaking, roman2014asymptotic}, we design our \textit{Multilevel Importance Sampling (MLIS)} module in the graph wavelet domain, which is responsible for subsampling the graph wavelet coefficients from observed data.

Given the wavelet coefficient matrix $\mathbf{W}\in\mathbb{R}^{N\times F}$ and a sampling budget $m=cN$, MLIS exploits the multilevel energy structure to allocate the budget across scales and select the most informative coefficients within each scale. Throughout this section, $W_{i,k}$ denotes the $(i,k)$ entry of
$\mathbf{W}\in\mathbb{R}^{N\times F}$ (node $i$, scale $k$). The energy of scale $k\in\{1,\dots,F\}$ is
\begin{equation}
    E_k = \sum_{i=1}^N W_{i,k}^2,
    \label{eq:scale_energy}
\end{equation}
and the fraction of the total budget allocated to scale $k$ is
\begin{equation}
    r_k = \frac{E_k}{\sum_{k'=1}^{F} E_{k'}}, \qquad
    m_k = \lfloor r_k \cdot m \rfloor,
    \label{eq:budget_alloc}
\end{equation}
so that $\sum_{k=1}^F m_k = m$ and scales with higher energy receive more
samples. For importance-weighted intra-scale sampling within scale $k$, we draw $m_k$ node indices without replacement according to the probability,
\begin{equation}
    p_{i,k} = \frac{W_{i,k}^2}{\sum_{i'=1}^{N} W_{i',k}^2},
    \label{eq:sampling_dist}
\end{equation}
concentrating measurements on nodes with larger-magnitude coefficients at
that scale.
Intuitively, dropping a large-magnitude coefficient increases the
reconstruction error more than dropping a near-zero one, so retaining
high-energy nodes is a natural proxy for minimizing reconstruction error
under a fixed budget.
The procedure yields a binary mask $\mathbf{S}\in\{0,1\}^{N\times F}$,
where $S_{i,k}=1$ if coefficient $(i,k)$ is selected, with $\|\mathbf{S}\|_0=m$. The sparse coefficient matrix is then obtained as
\begin{equation}
    \tilde{\mathbf{W}} = \mathbf{S} \odot \mathbf{W} \in \mathbb{R}^{N\times F},
    \label{eq:sparse_coeffs}
\end{equation}
where $\odot$ is the Hadamard product.
The pair $(\tilde{\mathbf{W}},\mathbf{S})$ is the compressed representation fed to the reconstruction model.

\subsection{Neural Inverse Graph Wavelet Transform (NIGWT).}
\label{subsec:nigwt}

NIGWT reconstructs $\hat{\mathbf{x}}\in\mathbb{R}^N$ from
$(\tilde{\mathbf{W}},\mathbf{S},\mathcal{G})$ using an Encode-Process-Decode architecture~\cite{sanchez2020learning,battaglia2018relational}, implementing learned scale embeddings and a residual IGWT connection. Let $D$ denote the hidden dimension and $\sigma$ the SiLU activation.

\textbf{Encoder.}
The Encoder network computes an initial node embedding matrix by combining sparse
graph wavelet coefficients $\tilde{\mathbf{W}}$, sparsification mask $\mathbf{S}$,
and learnable scale encodings $\mathbf{E}$.
Specifically, we first compute the encoded inputs by transforming the concatenation
of the sparse graph wavelet coefficients and the selection mask,
$\tilde{\mathbf{W}} \| \mathbf{S} \in \mathbb{R}^{N \times 2F}$:
\begin{equation}
    \mathbf{Z}^{\mathrm{input}} = \sigma\!\left(f^{(1)}_{\mathsf{Enc}}\!\left(\tilde{\mathbf{W}} \| \mathbf{S}\right)\right),
    \label{eq:enc_step1}
\end{equation}
where $f^{(1)}_{\mathsf{Enc}}: \mathbb{R}^{2F} \rightarrow \mathbb{R}^{D}$ is an MLP.
The main element of the Encoder is the learned scale-encodings.
Similar to positional encodings in the Transformer architecture~\cite{vaswani2017attention},
which provide tokens with an explicit notion of position, our learned scale-encodings
give each graph wavelet scale an explicit, learnable identity that the network can use
to distinguish coarse and fine scales.
Specifically, we learn a set of scale-encodings $\mathbf{e}_1, \dots, \mathbf{e}_F \in \mathbb{R}^D$
that act as basis vectors for scale information; stacking them gives
$\mathbf{E} \in \mathbb{R}^{F \times D}$.
For each node $i$ we compute the node-scale embedding
$\mathbf{z}^{\mathrm{scale}}_i \in \mathbb{R}^D$ by a weighted sum over scales:
\begin{equation}
    \mathbf{z}^{\mathrm{scale}}_i = \sum_{k=1}^F \tilde{W}_{i,k}\, \mathbf{e}_k,
    \quad\text{i.e.,}\quad
    \mathbf{Z}^{\mathrm{scale}}=\tilde{\mathbf{W}}\mathbf{E} \in \mathbb{R}^{N \times D},
    \label{eq:scale_embed}
\end{equation}
using the retained coefficient magnitude $\tilde{W}_{i,k}$ as the weight,
so that $\mathbf{z}^{\mathrm{scale}}_i$ encodes how much energy node $i$
carries at each scale.
Finally, we concatenate the node input and scale embedding matrices and apply the
last transformation to get the initial node embeddings:
\begin{equation}
    \mathbf{Z}^{(0)}= \sigma\!\left(f^{(2)}_{\mathsf{Enc}}\!\left( \mathbf{Z}^{\mathrm{input}} \| \mathbf{Z}^{\mathrm{scale}} \right)\right),
    \quad \mathbf{Z}^{(0)}\in\mathbb{R}^{N\times D},
    \label{eq:enc_step2}
\end{equation}
where $f^{(2)}_{\mathsf{Enc}}: \mathbb{R}^{2D} \rightarrow \mathbb{R}^{D}$ is an MLP.

\textbf{Processor.}
The Processor network comprises $L$ message-passing layers that process
the initial node embedding matrix and produces informative node representations for signal reconstruction.
In our implementation, each layer $f_{\mathsf{Proc}}^{(\ell)}$ is a GraphSAGE convolution (SAGEConv)~\cite{hamilton2017inductive} preceded by layer normalization $\mathrm{LN}(\cdot)$ and skip connection:
\begin{equation}
    \mathbf{Z}^{(\ell + 1)} = f_{\mathsf{Proc}}^{(\ell)}\!\left(\sigma\!\bigl(\mathrm{LN}\bigl(\mathbf{Z}^{(\ell)}\bigr)\bigr), \mathcal{G} \right) + \mathbf{Z}^{(\ell)},
    \quad \ell=0, \dots, L-1.
    \label{eq:processor}
\end{equation}

\textbf{Decoder.}
The final component transforms node representations to the reconstructed signal. This module learns the residual error between the original graph signal and its reconstruction obtained via the pseudo-inverse graph wavelet transform:
\begin{equation}
    \hat{\mathbf{x}} = \mathrm{IGWT}(\tilde{\mathbf{W}}) + f_{\mathsf{Dec}}\!\bigl(\mathbf{Z}^{(L)}\bigr),
    \label{eq:decoder}
\end{equation}
where $\mathrm{IGWT}(\tilde{\mathbf{W}})=\mathbf{\Psi}^\dagger\mathrm{vec}(\tilde{\mathbf{W}})\in\mathbb{R}^N$
is the Chebyshev-approximated inverse SGWT (computed without gradient updates), and $f_{\mathsf{Dec}}:\mathbb{R}^D\to\mathbb{R}$ is an MLP applied node-wise. The residual formulation allows NIGWT to refine an initial IGWT-based reconstruction, leading to improved optimization stability and faster convergence.

\subsection{Sparse Graph Autoencoder (SGAE).}
\label{subsec:sgae}

\begin{figure}[h]
  \centering
  \includegraphics[width=0.60\textwidth]{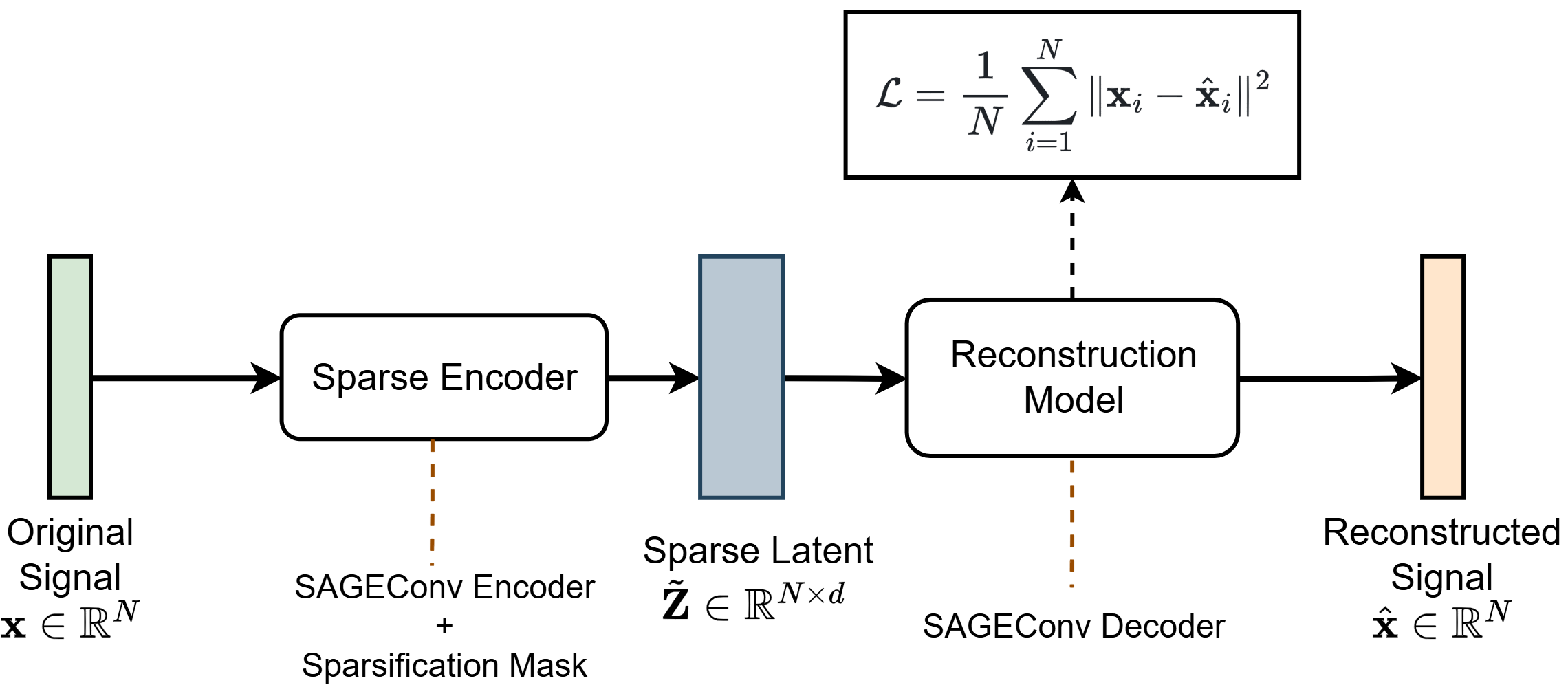}
  \caption{%
  Overview of the SGAE Framework.
  The raw signal $\mathbf{x}\in\mathbb{R}^{N}$ is encoded directly by the
 Sparse Encoder into a sparse latent ${\tilde{\mathbf{Z}}}\in\mathbb{R}^{N\times D_{\mathrm{SGAE}}}$ ($D_{\mathrm{SGAE}}=128$; $cN$ entries active). The Reconstruction Model (Graph Decoder) maps $\tilde{\mathbf{Z}}$ back to the signal domain.}
  \label{fig:sgae_architecture}
\vspace{-.12in}
\end{figure}

To isolate the contribution of the wavelet domain, we introduce the Sparse Graph Autoencoder (SGAE) as a baseline that operates directly on the raw signal without any SGWT preprocessing or multilevel sampling, shown in Figure~\ref{fig:sgae_architecture}.

\textbf{Sparse Encoder (Mask Generation Model).}
A stack of $L$ SAGEConv layers encodes the raw signal $\mathbf{x}\in\mathbb{R}^{N}$ into a dense latent $\mathbf{Z}\in\mathbb{R}^{N\times D_{\mathrm{SGAE}}}$. A differentiable sparse bottleneck then retains $m=cN$ latent entries via Straight-Through Estimation (STE) with temperature annealing:
\begin{equation}
    \tilde{\mathbf{Z}} = \mathrm{SparseBottleneck}(\mathbf{Z},\,\rho),
    \quad \|\tilde{\mathbf{Z}}\|_0 =cN,
    \label{eq:sgae_sparse}
\end{equation}
where $c=5\%$ is the same target compression ratio used for GWCS.

\textbf{Graph Decoder (Reconstruction Model).}
A symmetric stack of SAGEConv layers maps $\tilde{\mathbf{Z}}$ back to the
signal domain:
$\hat{\mathbf{x}}=\mathrm{Decoder}(\tilde{\mathbf{Z}},\mathcal{E})$. Crucially, SGAE learns a purely data-driven compressed representation as it bypasses the SGWT entirely and has no inductive bias towards wavelet sparsity.

\paragraph{Training Objectives.}

Both GWCS and SGAE are trained end-to-end by minimizing the same MSE reconstruction loss over the dataset with the Stochastic Gradient Descent (SGD). For GWCS, let $\theta$ denote the parameters of the NIGWT module. The reconstructed signal is given by
$\hat{\mathbf{x}}^{(j)}(\theta) := \mathrm{NIGWT}_{\theta}\!\left(\mathrm{MLIS}(\mathbf{W}), \mathcal{G} \right)$, and the MSE objective is
\begin{equation}
\theta^\star = \arg\min_{\theta}
\frac{1}{M}\sum_{j=1}^M \frac{1}{N^{(j)}}
\big\|\mathbf{x}^{(j)} - \hat{\mathbf{x}}^{(j)}(\theta)\big\|_2^2.
\end{equation}
SGAE is trained with the identical objective over its encoder-decoder parameters, with gradients passed through the sparse bottleneck via STE.

\begin{table*}[t]
\centering
\caption{Reconstruction performance on the ABL dataset. Our method achieves superior performance while using significantly fewer samples
(1-10\% of wavelet coefficients with $F=5$ filters and a single-channel signal, at $c=5-50\%$ of nodes). All GSS baselines use oracle bandwidth knowledge and optimized hyperparameters. Best results per column in bold.}
\label{tab:abl_results}
\small
\resizebox{\textwidth}{!}{
\begin{tabular}{@{}lc|cc|cc|cc@{}}
\toprule
& \textbf{Compression} & \multicolumn{2}{c|}{\textbf{Low ($c=5\%$)}} & \multicolumn{2}{c|}{\textbf{Medium ($c=25\%$)}} & \multicolumn{2}{c}{\textbf{High ($c=50\%$)}} \\
& \textbf{Ratio $c$} & \multicolumn{2}{c|}{$=1\%$ coeffs} & \multicolumn{2}{c|}{$=5\%$ coeffs} & \multicolumn{2}{c}{$=10\%$ coeffs} \\
\cmidrule(lr){3-4} \cmidrule(lr){5-6} \cmidrule(lr){7-8}
\textbf{Method} & & \textbf{MSE(x)} & \textbf{MSE(clean)} & \textbf{MSE(x)} & \textbf{MSE(clean)} & \textbf{MSE(x)} & \textbf{MSE(clean)} \\
\midrule
ESS & 5\% / 25\% / 50\% nodes & 0.18±0.16 & 0.14±0.15 & 0.05±0.03 & 0.02±0.02 & 0.039±0.027 & \textbf{0.007±0.007} \\
RSBS & 5\% / 25\% / 50\% nodes & 0.25±0.19 & 0.21±0.18 & 0.13±0.11 & 0.10±0.10 & 0.078±0.066 & 0.064±0.060 \\
FastGSSS & 5\% nodes only & 0.19±0.16 & 0.15±0.15 & --- & --- & --- & --- \\
BSGDA & 5\% / 25\% / 50\% nodes & 0.21±0.17 & 0.18±0.16 & 0.22±0.18 & 0.20±0.17 & 0.128±0.112 & 0.132±0.111 \\
\cmidrule{1-8}
SGAE & same \# latent coeffs & 0.194±0.141 & 0.163±0.126 & 0.105±0.069 & 0.105±0.058 & 0.027±0.020 & 0.049±0.025 \\
\midrule
\textbf{GWCS (Ours)} & 1\% / 5\% / 10\% coeffs & \textbf{0.13±0.10} & \textbf{0.09±0.09} & \textbf{0.04±0.02} & \textbf{0.019±0.019} & \textbf{0.021±0.014} & 0.013±0.009 \\
\bottomrule
\end{tabular}}
\vspace{-.12in}
\end{table*}

\begin{table*}[h]
\centering
\caption{Reconstruction RMSE (physical units, mean\,$\pm$\,std) on four PDE datasets at compression ratio $c=5\%$. \textbf{Bold}: best per dataset; \underline{underline}: second-best. $^\ddagger$Classical GSS values are
computed over 50 test samples, as full evaluation is computationally prohibitive (${\approx}$5--6\,h per sample). Supplementary MSE and relative $\ell_2$ metrics are in Appendix~\ref{appendix:supp_metrics}.}
\label{tab:pde_results}
\small
\setlength{\tabcolsep}{4pt}
\begin{tabular}{lcccc}
\toprule
\textbf{Method} &
  \textbf{Viscoelastic} &
  \textbf{Turbulent} &
  \textbf{Kolmogorov} &
  \textbf{Dynamic Stall} \\
\midrule
FastGSSS$^\ddagger$ & $0.0051\pm0.0010$ & $40.5\pm6.9$  & $0.834\pm0.110$ & $9310\pm1908$ \\
BSGDA$^\ddagger$    & $0.0040\pm0.0006$ & $8.9\pm1.5$   & $0.508\pm0.070$ & $5368\pm1040$ \\
RSBS$^\ddagger$     & $0.0044\pm0.0007$ & $9.2\pm1.6$   & $0.563\pm0.082$ & $6092\pm1203$ \\
GXN
  & $0.0052\pm0.0040$
  & $41.5\pm1.5$
  & $0.851\pm0.113$
  & $9350\pm1861$ \\
SGAE-50K
  & \underline{$0.0035\pm0.0026$}
  & \underline{$5.72\pm2.71$}
  & $\mathbf{0.287\pm0.059}$
  & \underline{$3211\pm819$} \\
SGAE-99K
  & $0.0037\pm0.0030$
  & $\mathbf{4.78\pm2.26}$
  & $0.328\pm0.050$
  & $\mathbf{2480\pm744}$ \\
\midrule
\textbf{GWCS (Ours)}
  & $\mathbf{0.0016\pm0.0011}$
  & $7.06\pm1.09$
  & \underline{$0.316\pm0.041$}
  & $9119\pm1700$ \\
\bottomrule
\end{tabular}
\vspace{-.12in}
\end{table*}


\section{Experiments}
\label{experiments}

\subsection{Datasets.}
\label{subsec:datasets}

\textbf{Synthetic Approximately Band-limited (ABL) signals.}
We employ a class of ABL graph signals with additive Gaussian noise, modeled as $\mathbf{x} = \mathbf{U}_R\mathbf{a} + \mathbf{n}$, where $\mathbf{U}_R \in \mathbb{R}^{N \times R}$ contains the $R$ lowest-frequency eigenvectors of the normalized Laplacian $\mathbf{L}$,
$\mathbf{a}\sim\mathcal{N}(\mathbf{0},\sigma^2\mathbf{I})$ is a vector of
band-limited coefficients, and $\mathbf{n}$ is additive Gaussian noise.
Signals consistent with this model have most of their energy in the
low-frequency components. We generate random graphs from the grid, Erd\H{o}s--R\'{e}nyi~\cite{erdds1959random},
and Barab\'{a}si--Albert~\cite{barabasi1999emergence} models with $N\in[100,625]$, and for each graph generate a random band-limited signal with additive noise. Details of the data-generating process are provided in Appendix~\ref{ABL}.

\textbf{Physics-based PDE Datasets.}
We further evaluate our model on four PDE datasets drawn from the WELL benchmark~\cite{ohana2024well} and the PBFM benchmark~\cite{pbfm2026}, both providing standardized, high-resolution PDE simulations for SciML.

\textit{Turbulent Radiative Layer 2D ($N=49{,}152$).}
This dataset simulates cold dense gas clumps moving through a surrounding hotter gas, mixing due to turbulence at their interface in astrophysical environments. The mixing creates an intermediate temperature phase that rapidly cools by radiative cooling, causing the mixed gas to join the cold phase. We compress and reconstruct the \textbf{density} field.

\textit{Viscoelastic Instability ($N=91{,}136$).}
This dataset captures the evolution of viscoelastic fluid flows exhibiting elastic instabilities, governed by coupled nonlinear PDEs describing the interaction between velocity and stress fields, with complex multiscale spatiotemporal dynamics. We reconstruct the \textbf{pressure} field.

\textit{Kolmogorov Flow ($N=16{,}384$).}
This dataset represents a canonical turbulent flow driven by a sinusoidal forcing, commonly used as a benchmark for studying transition to turbulence and chaotic dynamics, with rich vortex structures and multiscale behavior. We reconstruct the \textbf{velocity$_x$} field.

\textit{Dynamic Stall ($N=16{,}384$).}
This dataset models unsteady aerodynamic flows around an airfoil undergoing rapid changes in angle of attack, leading to flow separation and vortex shedding. The data captures highly nonlinear transient phenomena relevant to aerospace applications. We reconstruct the \textbf{pressure} field.

\subsection{Baselines.}
\label{subsec:baselines}
\textbf{Graph signal sampling (GSS).}
We compare against four classical non-parametric GSS algorithms that sequentially select optimal node subsets and provide theoretical reconstruction guarantees: ESS~\cite{ESS}, RSBS~\cite{RSBS}, FastGSSS~\cite{FastGSSS}, and BSGDA~\cite{BSGDA}, all implemented via the \texttt{THGSP} package~\cite{thgsp}\footnote{\url{https://github.com/bwdeng20/thgsp}}. To ensure a fair comparison, we conduct comprehensive hyperparameter optimization for all GSS methods on the validation split of the ABL dataset.
Since GSS methods are non-parametric and operate on individual graph signals without training, we use grid search over method-specific hyperparameter spaces to minimize validation mean-squared error (MSE).
Details of the hyperparameter search of GSS methods can be found in Appendix \ref{gss-optim}. Additionally, we provide GSS methods with oracle knowledge of the true signal bandwidth $R$ that represents an optimistic baseline.

\textbf{GXN.}
A joint neural sampling-and-recovery baseline: NeuralSampling selects
$=5\%$ of nodes via mutual information-based node scoring; NeuralRecovery reconstructs the full signal via a graph polynomial network~\cite{chen2020sampling}.

\textbf{SGAE.}
Our sparse graph autoencoder (Section~\ref{subsec:sgae}), is evaluated at two parameter budgets: \textbf{SGAE-99K} (${\approx}99{,}000$ parameters, latent dimension $D_{\mathrm{SGAE}}=128$, $L=4$ SAGEConv encoder/decoder layers) and \textbf{SGAE-50K} (${\approx}50{,}000$ parameters), both at target compression ratio $c=5\%$ matched to GWCS.

\textbf{GWCS.}
Our proposed model uses hidden dimension $D=256$, $L=5$ SAGEConv layers,
$F=5$ wavelet filters ($J=4$ bandpass plus one lowpass), Chebyshev degree
$K=100$, yielding ${\approx}993{,}000$ parameters. A lightweight variant ($D=96$, $L=3$, ${\approx}99{,}000$ parameters) is used for parameter-matched comparison against SGAE-99K. All models use a compression ratio $c=5\%$.

\subsection{Evaluation.}
\label{subsec:evaluation}

We report the Root Mean Squared Error (RMSE) in physical units, computed after de-normalizing predictions to the original signal scale. The sample-level RMSE for sample $j$ is given by
$\mathrm{RMSE}^{(j)}=\|\hat{\mathbf{x}}^{(j)}-\mathbf{x}^{(j)}\|_2/\sqrt{N}$, and we compute the mean and standard deviation across all test samples.

\subsection{Results on ABL Signals.}
\label{subsec:results_abl}

Table~\ref{tab:abl_results} reports MSE across three compression levels.
GWCS outperforms all GSS baselines (ESS, RSBS, FastGSSS, and BSGDA) at the low and medium budgets by a large margin \emph{without} oracle bandwidth knowledge, and matches the best GSS method (ESS) at the high budget. SGAE also underperforms GWCS at all budgets on ABL signals, confirming that the wavelet-domain representation provides a strong inductive bias for approximately band-limited data.
Figures~\ref{fig:app_grid_1}--\ref{fig:app_er_3} in Appendix~\ref{appendix:qual_all} further visualize the SGWT coefficients, MLIS sparsification mask, and NIGWT reconstruction for four representative ABL test samples over grid and Erd\H{o}s--R\'{e}nyi graphs.

\subsection{Results on PDE Datasets.}
\label{subsec:results_pde}

\begin{figure*}[p]
  \centering
  \begin{minipage}[c]{0.13\textwidth}\small\raggedright
    \textbf{(a)} \mbox{GWCS}\\[2pt]\mbox{RMSE\,$=\,6.90$}
  \end{minipage}%
  \hspace{1pt}%
  \begin{minipage}[c]{0.75\textwidth}
    \includegraphics[width=\linewidth]{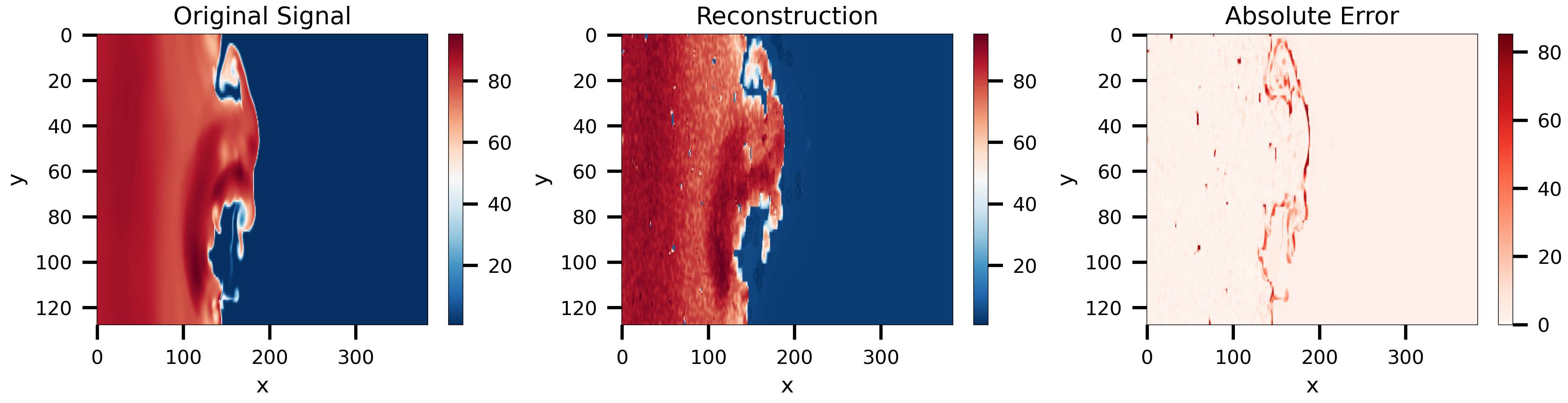}
  \end{minipage}\\[3pt]
  \begin{minipage}[c]{0.13\textwidth}\small\raggedright
    \textbf{(b)} \mbox{SGAE-99K}\\[2pt]\mbox{RMSE\,$=\,9.32$}
  \end{minipage}%
  \hspace{1pt}%
  \begin{minipage}[c]{0.75\textwidth}
    \includegraphics[width=\linewidth]{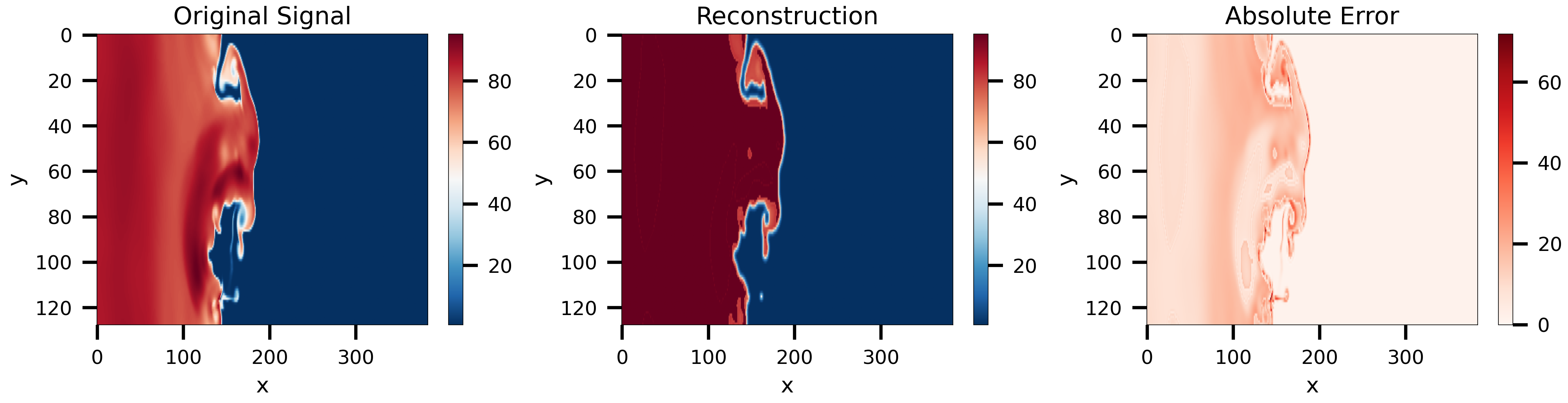}
  \end{minipage}\\[3pt]
  \begin{minipage}[c]{0.13\textwidth}\small\raggedright
    \textbf{(c)} \mbox{SGAE-50K}\\[2pt]\mbox{RMSE\,$=\,10.33$}
  \end{minipage}%
  \hspace{1pt}%
  \begin{minipage}[c]{0.75\textwidth}
    \includegraphics[width=\linewidth]{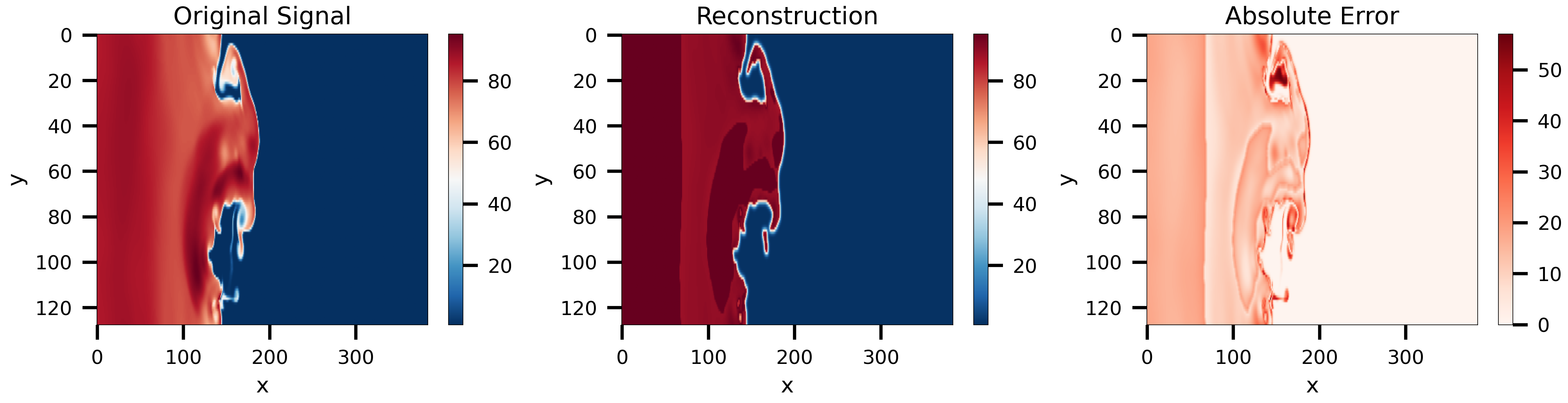}
  \end{minipage}\\[3pt]
  \begin{minipage}[c]{0.13\textwidth}\small\raggedright
    \textbf{(d)} \mbox{GXN}\\[2pt]\mbox{RMSE\,$=\,38.55$}
  \end{minipage}%
  \hspace{1pt}%
  \begin{minipage}[c]{0.75\textwidth}
    \includegraphics[width=\linewidth]{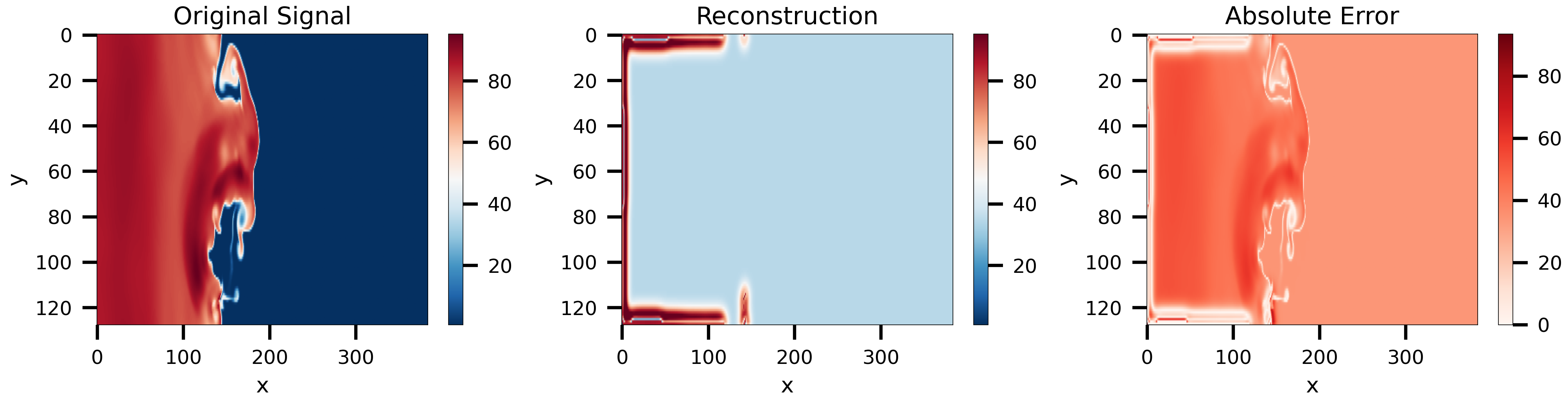}
  \end{minipage}
  \caption{Qualitative reconstruction of the Turbulent Radiative Layer density field
  (test sample~48, compression ratio $c=5\%$). Columns: ground truth, reconstruction, absolute error.}
    \label{fig:turbulent_qualitative}
  \vspace{-.12in}
\end{figure*}

Reconstruction performance across the four PDE simulation datasets, measured by RMSE, is shown in Table~\ref{tab:pde_results}. The central finding is that, GWCS excels on spectrally structured PDE fields and is competitive on moderate turbulence, while SGAE is preferable when energy is distributed broadly across wavelet scales. Supplementary quantitative metrics (MSE and relative $\ell_2$ error) for all neural models are reported in Appendix~\ref{appendix:supp_metrics}, confirming the same ordering across metrics. This pattern is entirely consistent with the theoretical motivation of GWCS: wavelet-domain sampling provides a decisive advantage precisely when the signal's energy is concentrated in a small number of scales. Figure~\ref{fig:turbulent_qualitative} shows qualitative reconstruction of the Turbulent Radiative Layer density field on a representative test sample (sample~48) at compression ratio $c=5\%$. Figure~\ref{fig:qual_turbulent}(a) shows that GWCS recovers the dominant hot--cold density interface: the sharp boundary between the cold dense phase and the surrounding hot gas is well preserved, and the residual absolute error is concentrated along the turbulent mixing layer boundary where broadband fluctuations exceed the capacity of the fixed five-filter SGWT at $c=5\%$. Figures~\ref{fig:turbulent_qualitative}(b) and (c) show that SGAE-99K and SGAE-50K achieve lower per-sample RMSE; their unconstrained, data-driven latent representations capture the broadband energy components that GWCS's fixed wavelet filterbank discards. The larger capacity SGAE-99K (${\approx}99$K parameters) outperforms SGAE-50K (${\approx}50$K) on this dataset, confirming that representational capacity matters for broadband flows. Figure~\ref{fig:turbulent_qualitative}(d) illustrates the failure of GXN: the reconstruction is nearly uniform, failing entirely to resolve the density contrast, consistent with its mean RMSE of $41.49$ across the full test set. The root cause is that turbulent radiative layer flows have broad-band energy distributed across all wavelet scales; the strict 1\% wavelet coefficient budget in GWCS cannot retain enough energy in all scales simultaneously, whereas SGAE's learned latent space is unconstrained and can allocate capacity freely across the signal's energy distribution. Qualitative reconstructions for all four datasets and baselines are provided in Appendix~\ref{appendix:qual_all}.

\begin{figure*}[p]
  \centering
  \subfigure[Viscoelastic Instability: per-sample RMSE (all test samples; neural models only)]{%
    \includegraphics[width=0.49\textwidth]{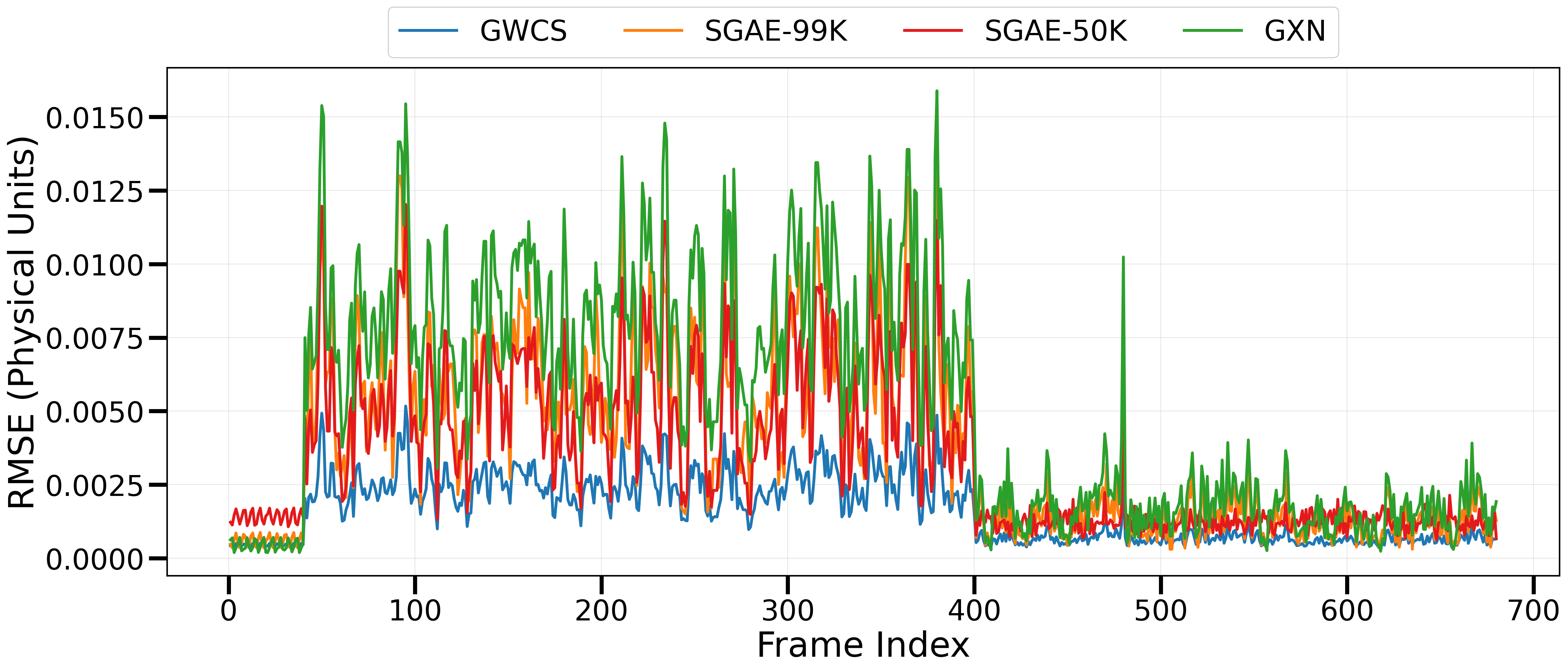}}%
  \hfill
  \subfigure[Kolmogorov Flow: per-sample RMSE (first 50 test samples; neural\,+\,classical baselines)]{%
    \includegraphics[width=0.49\textwidth]{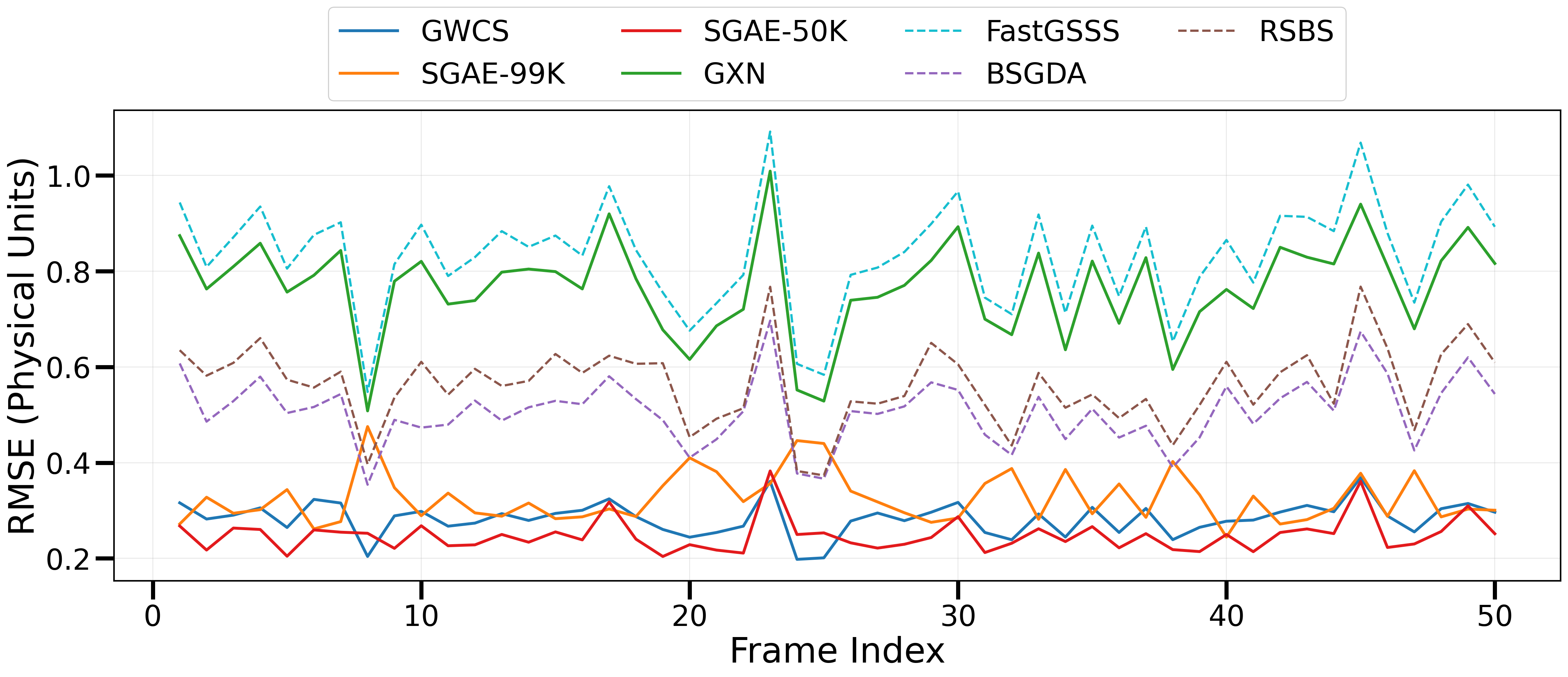}}
  \caption{%
  Per-sample RMSE trajectories (physical units).
  \textbf{(a) Viscoelastic Instability} (all test samples, neural models only):
  GWCS maintains the lowest RMSE throughout the test set.
  \textbf{(b) Kolmogorov Flow} (first 50 samples, neural and classical):
  all three sparsity-based models cluster well below GXN and classical GSS baseline (FastGSSS, RSBS, BSGDA),
  confirming the necessity of learned representations. 
  }
  \label{fig:rmse_curves}
  \vspace{-.12in}
\end{figure*}

Figure~\ref{fig:rmse_curves} shows per-sample RMSE trajectories for two representative cases. Panel~(a) covers all test samples for Viscoelastic Instability using neural models only; classical GSS baselines are omitted because their per-sample cost is computationally expensive on this large mesh ($N{=}91{,}136$ nodes). Panel~(b) shows the first 50 Kolmogorov test samples with both neural and classical baselines; the truncation to 50 samples avoids visual crowding with eight overlapping series and matches the subset for which classical methods were evaluated. For Viscoelastic Instability, GWCS achieves the lowest RMSE ($0.0016$), outperforming SGAE-50K ($\times2.2$ better), SGAE-99K ($\times2.3$), and GXN ($\times3.2$). Viscoelastic flows have spectrally structured spatial patterns governed by smooth PDEs; the SGWT captures this structure efficiently and GWCS's scale-aware encoder and residual IGWT connection leverage it strongly. For Kolmogorov Flow, SGAE-50K achieves the lowest RMSE ($0.287$), with GWCS ($0.319$) and SGAE-99K ($0.328$) close behind; GXN substantially underperforms ($0.851$). Notably, the smaller SGAE-50K outperforms the larger SGAE-99K on this dataset, suggesting that the moderate-turbulence flow does not require the additional representational capacity and that the 99K variant may be slightly overfit. As shown in Figure~\ref{fig:rmse_curves}(b), all three neural methods cluster well below the classical GSS baselines (FastGSSS, RSBS, BSGDA, RMSE $\approx\!0.6$--$1.1$), confirming the value of learned representations for turbulent flows. GWCS is competitive with SGAE despite its structured wavelet prior operating at the same $c=5\%$ compression ratio.

For Turbulent Radiative Layer, SGAE-99K achieves the best mean RMSE ($4.78$), with GWCS second ($7.06$), SGAE-50K third ($5.72$), and GXN failing entirely ($41.5$). Turbulent flows have broad-band energy across all wavelet scales; the strict 1\% wavelet coefficient budget ($c=5\%$) in NIGWT discards energy that SGAE's data-driven latent representation retains. For Dynamic Stall dataset, SGAE-99K dominates in terms of lowest reconstruction error; whereas GWCS and GXN perform substantially poorly. Dynamic stall features sharp transient spatial gradients at the airfoil leading edge; the fixed 5-filter SGWT may not resolve these structures within a 1\% coefficient budget ($c=5\%$). Across all four datasets, sparsity methods (GWCS and SGAE) consistently outperform classical GSS and GXN, validating the learning-based compressed sensing approach. GWCS has a decisive advantage on spectrally regular PDE data (Viscoelastic), is competitive on moderate turbulence, and is outperformed by SGAE on highly complex or transient datasets. This pattern is consistent with the theoretical motivation: wavelet-domain sampling benefits signals whose energy is concentrated in a small number of scales and spatial locations.

\subsection{Discussion.}
\label{discussion}
GWCS supports both \emph{transductive} and \emph{inductive} modes of use.
In the transductive setting, GWCS compresses an entire dataset for storage
or transmission at a given compression ratio, producing a sparse
wavelet-domain representation of the dataset along with a trained model
that can later reconstruct the full signals.
In the inductive setting, GWCS is trained once and then applied to unseen data as a sampler and
reconstructor for downstream machine learning. MLIS compresses new test
signals into sparse graph wavelet coefficients that downstream processing
can consume directly, and NIGWT recovers the original domain signal only when it is needed.

GWCS differs from classical CS along two axes. Classical CS is posed
\emph{online}, with sampling designed at acquisition time, whereas GWCS is
\emph{offline}, operating on signals that are already fully available.
Classical CS also reconstructs each sample by solving an $\ell_1$-regularized
optimization problem with no learning capacity, whereas GWCS trains a
neural model to reconstruct signals from their retained coefficients.
The online setting in CS applications is motivated by the cost of physical measurements, a cost
that is largely irrelevant in SciML, where data
typically consists of large volumes of simulated PDE solutions already
available offline. This offline setting instead incurs the cost of
storing the full signal and its coefficients, and the computational cost
of the Chebyshev-approximated SGWT. In exchange, once trained,
reconstruction requires only a single MLIS pass and a forward pass of
NIGWT.

MLIS is a heuristic sampler that exploits the sparsity of natural signals
in the wavelet domain together with multilevel, energy-based coefficient
retention, rather than learning which coefficients to keep. Compared to
learning-based samplers such as GXN and SGAE, this yields an interpretable
compressed representation and a more stable training process for the
reconstruction model.
This heuristic performs best when a signal's energy
is concentrated in a few wavelet scales, as for Viscoelastic Instability,
where GWCS achieves its best reconstruction among the four PDE datasets.
When energy is instead distributed broadly across scales,
as in turbulent or transient flows,
a fixed wavelet basis has less structure to exploit,
which may account for the narrower margin between GWCS and SGAE
observed on these datasets.

As future work, we plan to incorporate the temporal dimension of PDE
signals into the MLIS and NIGWT mechanisms, and to explore downstream
SciML tasks performed directly in the sparse graph
wavelet domain.

\section{Conclusion}
\label{conclusion}
We introduced Graph Wavelet Compressed Sensing (GWCS), a learning-based
framework for compressing graph-structured scientific data, inspired by
compressed sensing theory.
Its core contributions are the nonparametric Multilevel Importance Sampling (MLIS) module and
the scale-aware Neural Inverse Graph Wavelet Transform (NIGWT) recovery model.
For controlled comparison, we additionally introduced a Sparse Graph Autoencoder (SGAE),
a fully trainable, wavelet-free baseline.
GWCS supports both transductive and inductive settings.
It can compress an entire dataset for storage or transmission,
or provide interpretable sparse representations for downstream machine learning tasks.
Experimental results demonstrate the effectiveness of
GWCS in compressing and reconstructing signals compared to both
classical graph signal sampling and recent learning-based neural
architectures.

\section*{Acknowledgments.}
This work was supported by the DOE SEA-CROGS project (DE-SC0023191), AFOSR project (FA9550-24-1-0231). We also thank the computing resources provided by the High Performance Computing (HPC) facility at NJIT.

\bibliographystyle{siamplain}
\bibliography{ref}

@article{kovachki2023neural,
  title={Neural operator: Learning maps between function spaces with applications to pdes},
  author={Kovachki, Nikola and Li, Zongyi and Liu, Burigede and Azizzadenesheli, Kamyar and Bhattacharya, Kaushik and Stuart, Andrew and Anandkumar, Anima},
  journal={Journal of Machine Learning Research},
  volume={24},
  number={89},
  pages={1--97},
  year={2023}
}

@article{raissi2019physics,
  title={Physics-informed neural networks: A deep learning framework for solving forward and inverse problems involving nonlinear partial differential equations},
  author={Raissi, Maziar and Perdikaris, Paris and Karniadakis, George E},
  journal={Journal of Computational physics},
  volume={378},
  pages={686--707},
  year={2019},
  publisher={Elsevier}
}

@article{candes2006robust,
  title={Robust uncertainty principles: Exact signal reconstruction from highly incomplete frequency information},
  author={Cand{\`e}s, Emmanuel J and Romberg, Justin and Tao, Terence},
  journal={IEEE Transactions on information theory},
  volume={52},
  number={2},
  pages={489--509},
  year={2006},
  publisher={IEEE}
}

@article{donoho2006compressed,
  title={Compressed sensing},
  author={Donoho, David L},
  journal={IEEE Transactions on information theory},
  volume={52},
  number={4},
  pages={1289--1306},
  year={2006},
  publisher={IEEE}
}

@inproceedings{adcock2017breaking,
  title={Breaking the coherence barrier: A new theory for compressed sensing},
  author={Adcock, Ben and Hansen, Anders C and Poon, Clarice and Roman, Bogdan},
  booktitle={Forum of mathematics, sigma},
  volume={5},
  pages={e4},
  year={2017},
  organization={Cambridge University Press}
}

@article{roman2014asymptotic,
  title={On asymptotic structure in compressed sensing},
  author={Roman, Bogdan and Hansen, Anders and Adcock, Ben},
  journal={arXiv preprint arXiv:1406.4178},
  year={2014}
}

@inproceedings{adcock2015quest,
  title={The quest for optimal sampling: Computationally efficient, structure-exploiting measurements for compressed sensing},
  author={Adcock, Ben and Hansen, Anders C and Roman, Bogdan},
  booktitle={Compressed Sensing and its Applications: MATHEON Workshop 2013},
  pages={143--167},
  year={2015},
  organization={Springer}
}

@article{shuman2013emerging,
  title={The emerging field of signal processing on graphs: Extending high-dimensional data analysis to networks and other irregular domains},
  author={Shuman, David I and Narang, Sunil K and Frossard, Pascal and Ortega, Antonio and Vandergheynst, Pierre},
  journal={IEEE signal processing magazine},
  volume={30},
  number={3},
  pages={83--98},
  year={2013},
  publisher={IEEE}
}

@article{hammond2011wavelets,
  title={Wavelets on graphs via spectral graph theory},
  author={Hammond, David K and Vandergheynst, Pierre and Gribonval, R{\'e}mi},
  journal={Applied and Computational Harmonic Analysis},
  volume={30},
  number={2},
  pages={129--150},
  year={2011},
  publisher={Elsevier}
}

@inproceedings{ricaud2013sparsity,
  title={On the sparsity of wavelet coefficients for signals on graphs},
  author={Ricaud, Benjamin and Shuman, David I and Vandergheynst, Pierre},
  booktitle={Wavelets and Sparsity XV},
  volume={8858},
  pages={422--428},
  year={2013},
  organization={SPIE}
}

@misc{thgsp,
  author = {Bowen Deng},
  title = {ThGSP: A PyTorch-based Graph Signal Processing Library},
  year = {2021},
  publisher = {GitHub},
  journal = {GitHub repository},
  howpublished = {\url{https://github.com/bwdeng20/thgsp}},
}

@article{ESS,
  title={Efficient sampling set selection for bandlimited graph signals using graph spectral proxies},
  author={Anis, Aamir and Gadde, Akshay and Ortega, Antonio},
  journal={IEEE Transactions on Signal Processing},
  volume={64},
  number={14},
  pages={3775--3789},
  year={2016},
  publisher={IEEE}
}

@article{RSBS,
  title={Random sampling of bandlimited signals on graphs},
  author={Puy, Gilles and Tremblay, Nicolas and Gribonval, R{\'e}mi and Vandergheynst, Pierre},
  journal={Applied and Computational Harmonic Analysis},
  volume={44},
  number={2},
  pages={446--475},
  year={2018},
  publisher={Elsevier}
}

@article{FastGSSS,
  title={Eigendecomposition-free sampling set selection for graph signals},
  author={Sakiyama, Akie and Tanaka, Yuichi and Tanaka, Toshihisa and Ortega, Antonio},
  journal={IEEE Transactions on Signal Processing},
  volume={67},
  number={10},
  pages={2679--2692},
  year={2019},
  publisher={IEEE}
}

@article{BSGDA,
  title={Fast graph sampling set selection using gershgorin disc alignment},
  author={Bai, Yuanchao and Wang, Fen and Cheung, Gene and Nakatsukasa, Yuji and Gao, Wen},
  journal={IEEE Transactions on signal processing},
  volume={68},
  pages={2419--2434},
  year={2020},
  publisher={IEEE}
}

@article{battaglia2018relational,
  title={Relational inductive biases, deep learning, and graph networks},
  author={Battaglia, Peter W and Hamrick, Jessica B and Bapst, Victor and Sanchez-Gonzalez, Alvaro and Zambaldi, Vinicius and Malinowski, Mateusz and Tacchetti, Andrea and Raposo, David and Santoro, Adam and Faulkner, Ryan and others},
  journal={arXiv preprint arXiv:1806.01261},
  year={2018}
}

@inproceedings{sanchez2020learning,
  title={Learning to simulate complex physics with graph networks},
  author={Sanchez-Gonzalez, Alvaro and Godwin, Jonathan and Pfaff, Tobias and Ying, Rex and Leskovec, Jure and Battaglia, Peter},
  booktitle={International conference on machine learning},
  pages={8459--8468},
  year={2020},
  organization={PMLR}
}

@article{kipf2016variational,
  title={Variational graph auto-encoders},
  author={Kipf, Thomas N and Welling, Max},
  journal={arXiv preprint arXiv:1611.07308},
  year={2016}
}

@article{vaswani2017attention,
  title={Attention is all you need},
  author={Vaswani, Ashish and Shazeer, Noam and Parmar, Niki and Uszkoreit, Jakob and Jones, Llion and Gomez, Aidan N and Kaiser, {\L}ukasz and Polosukhin, Illia},
  journal={Advances in neural information processing systems},
  volume={30},
  year={2017}
}

@article{ohana2024well,
  title={The well: a large-scale collection of diverse physics simulations for machine learning},
  author={Ohana, Ruben and McCabe, Michael and Meyer, Lucas and Morel, Rudy and Agocs, Fruzsina and Beneitez, Miguel and Berger, Marsha and Burkhart, Blakesly and Dalziel, Stuart and Fielding, Drummond and others},
  journal={Advances in Neural Information Processing Systems},
  volume={37},
  pages={44989--45037},
  year={2024}
}

@article{barabasi1999emergence,
  title={Emergence of scaling in random networks},
  author={Barab{\'a}si, Albert-L{\'a}szl{\'o} and Albert, R{\'e}ka},
  journal={science},
  volume={286},
  number={5439},
  pages={509--512},
  year={1999},
  publisher={American Association for the Advancement of Science}
}

@article{erdds1959random,
  title={On random graphs I},
  author={ERDdS, P and R\&wi, A},
  journal={Publ. math. debrecen},
  volume={6},
  number={290-297},
  pages={18},
  year={1959}
}

@article{lu2021learning,
  title={Learning nonlinear operators via DeepONet based on the universal approximation theorem of operators},
  author={Lu, Lu and Jin, Pengzhan and Pang, Guofei and Zhang, Zhongqiang and Karniadakis, George Em},
  journal={Nature machine intelligence},
  volume={3},
  number={3},
  pages={218--229},
  year={2021},
  publisher={Nature Publishing Group UK London}
}

@article{lustig2007sparse,
  title={Sparse MRI: The application of compressed sensing for rapid MR imaging},
  author={Lustig, Michael and Donoho, David and Pauly, John M},
  journal={Magnetic Resonance in Medicine: An Official Journal of the International Society for Magnetic Resonance in Medicine},
  volume={58},
  number={6},
  pages={1182--1195},
  year={2007},
  publisher={Wiley Online Library}
}

@article{chen2008prior,
  title={Prior image constrained compressed sensing (PICCS): a method to accurately reconstruct dynamic CT images from highly undersampled projection data sets},
  author={Chen, Guang-Hong and Tang, Jie and Leng, Shuai},
  journal={Medical physics},
  volume={35},
  number={2},
  pages={660--663},
  year={2008},
  publisher={Wiley Online Library}
}

@article{leary2013compressed,
  title={Compressed sensing electron tomography},
  author={Leary, Rowan and Saghi, Zineb and Midgley, Paul A and Holland, Daniel J},
  journal={Ultramicroscopy},
  volume={131},
  pages={70--91},
  year={2013},
  publisher={Elsevier}
}

@article{hamilton2017inductive,
  title={Inductive representation learning on large graphs},
  author={Hamilton, Will and Ying, Zhitao and Leskovec, Jure},
  journal={Advances in neural information processing systems},
  volume={30},
  year={2017}
}

@article{chen2020sampling,
  title={Sampling and recovery of graph signals based on graph neural networks},
  author={Chen, Siheng and Li, Maosen and Zhang, Ya},
  journal={arXiv preprint arXiv:2011.01412},
  year={2020}
}

@inproceedings{pbfm2026,
    title={Physics vs Distributions: Pareto Optimal Flow Matching with Physics Constraints},
    author={Giacomo Baldan and Qiang Liu and Alberto Guardone and Nils Thuerey},
    booktitle={The Fourteenth International Conference on Learning Representations},
    year={2026}
}

@inproceedings{kipf2017semi,
  title={Semi-Supervised Classification with Graph Convolutional Networks},
  author={Kipf, Thomas N. and Welling, Max},
  booktitle={International Conference on Learning Representations (ICLR)},
  year={2017}
}

@article{chen2021graph,
  title={Graph unrolling networks: Interpretable neural networks for graph signal denoising},
  author={Chen, Siheng and Eldar, Yonina C and Zhao, Lingxiao},
  journal={IEEE Transactions on Signal Processing},
  volume={69},
  pages={3699--3713},
  year={2021},
  publisher={IEEE}
}

@article{chen2015discrete,
  title={Discrete signal processing on graphs: Sampling theory<? pub \_newline=""?},
  author={Chen, Siheng and Varma, Rohan and Sandryhaila, Aliaksei and Kova{\v{c}}evi{\'c}, Jelena},
  journal={IEEE transactions on signal processing},
  volume={63},
  number={24},
  pages={6510--6523},
  year={2015},
  publisher={IEEE}
}

@article{anis2016efficient,
  title={Efficient sampling set selection for bandlimited graph signals using graph spectral proxies},
  author={Anis, Aamir and Gadde, Akshay and Ortega, Antonio},
  journal={IEEE Transactions on Signal Processing},
  volume={64},
  number={14},
  pages={3775--3789},
  year={2016},
  publisher={IEEE}
}

@article{marques2015sampling,
  title={Sampling of graph signals with successive local aggregations},
  author={Marques, Antonio G and Segarra, Santiago and Leus, Geert and Ribeiro, Alejandro},
  journal={IEEE Transactions on Signal Processing},
  volume={64},
  number={7},
  pages={1832--1843},
  year={2015},
  publisher={IEEE}
}

@inproceedings{tremblay2017graph,
  title={Graph sampling with determinantal processes},
  author={Tremblay, Nicolas and Amblard, Pierre-Olivier and Barthelm{\'e}, Simon},
  booktitle={2017 25th European signal processing conference (EUSIPCO)},
  pages={1674--1678},
  year={2017},
  organization={IEEE}
}

@inproceedings{gao2019graph,
  title={Graph u-nets},
  author={Gao, Hongyang and Ji, Shuiwang},
  booktitle={international conference on machine learning},
  pages={2083--2092},
  year={2019},
  organization={PMLR}
}

@article{schneider2010wavelet,
  title={Wavelet methods in computational fluid dynamics},
  author={Schneider, Kai and Vasilyev, Oleg V},
  journal={Annual review of fluid mechanics},
  volume={42},
  number={1},
  pages={473--503},
  year={2010},
  publisher={Annual Reviews}
}

@book{ortega2022introduction,
  title={Introduction to graph signal processing},
  author={Ortega, Antonio},
  year={2022},
  publisher={Cambridge University Press}
}

@article{nouranizadeh2021maximum,
  title={Maximum entropy weighted independent set pooling for graph neural networks},
  author={Nouranizadeh, Amirhossein and Matinkia, Mohammadjavad and Rahmati, Mohammad and Safabakhsh, Reza},
  journal={arXiv preprint arXiv:2107.01410},
  year={2021}
}

@article{ying2018hierarchical,
  title={Hierarchical graph representation learning with differentiable pooling},
  author={Ying, Zhitao and You, Jiaxuan and Morris, Christopher and Ren, Xiang and Hamilton, Will and Leskovec, Jure},
  journal={Advances in neural information processing systems},
  volume={31},
  year={2018}
}

\newpage

\appendix
\section{Chebyshev polynomial approximation of SGWT.}
\label{chebyshev}
Concretely, we map the spectral interval $[\lambda_{\min},\lambda_{\max}]$ affinely to $[-1,1]$ (with $a=(\lambda_{\max}-\lambda_{\min})/2$, $b=(\lambda_{\max}+\lambda_{\min})/2$) and approximate each kernel by a truncated Chebyshev series
\[
p(\lambda)=\tfrac{1}{2}c_0 + \sum_{k=1}^{K} c_k\,T_k\!\Big(\frac{\lambda-b}{a}\Big),
\]
where $T_0(y)=1$, $T_1(y)=y$ and $T_{k+1}(y)=2yT_k(y)-T_{k-1}(y)$. Replacing $\lambda$ by the matrix $L$ (via the shifted operator $\tilde L=(L-bI)/a$) yields the node-domain filter $p(L)$. The action of $p(L)$ on a signal $x$ is evaluated by the stable recurrence applied to vectors:
\begin{align*}
T_0(L)x&=x,\qquad T_1(L)x=\tilde L x, \\ \qquad T_{k+1}(L)x &=2\tilde L\,T_k(L)x - T_{k-1}(L)x,
\end{align*}
and accumulating
\[
p(L)x=\tfrac{1}{2}c_0 x + \sum_{k=1}^K c_k\,T_k(L)x.
\]
This avoids eigendecomposition and requires only repeated sparse matrix--vector products with $\tilde L$.

The Chebyshev coefficients $\{c_k\}$ are obtained by projecting the kernel onto the Chebyshev basis (equivalently via a discrete-cosine--type sampling/projection on standard Chebyshev nodes); coefficients can be computed up to a chosen degree and then truncated, so no per-degree optimization is required. Finally, if $K$ is the chosen polynomial degree and $|\mathcal E|$ the number of graph edges, the dominant cost is $K$ sparse mat--vecs, i.e.\ $O(K|\mathcal E|)$ (plus $O(N\sum_j M_j)$ to form all filter outputs when degrees $M_j$ differ across scales); memory overhead is modest (recurrence needs only a few $N$-vectors, practically $O(N)$).

\section{Synthetic data generating process}
\label{ABL}

\textbf{Graph Generation.}
We generate synthetic datasets using three random graph models with varying structural properties.
For grid graphs, we construct 2D lattice structures where each node is connected to its immediate neighbors, resulting in graphs with $N$ nodes arranged in rectangular grids.
For Erdős–Rényi graphs, we sample from $\mathcal{G}(N, p)$ where each possible edge is included independently with probability $p \in [0.01, 0.05]$, producing graphs with relatively homogeneous degree distributions.
For Barabási–Albert graphs, we use preferential attachment starting from a small initial graph, where each new node connects to $m \in \{3, 4, 5\}$ existing nodes with probability proportional to their degrees, resulting in scale-free networks with power-law degree distributions.

For each graph type, we generate $3,000$ instances with the number of nodes $N$ uniformly sampled from $[100, 625]$, providing diversity in graph sizes and topological characteristics.

\textbf{Signal generation.}
For each generated graph, the signal generation process proceeds as follows. First, we compute the normalized Laplacian matrix $\mathbf{L}_{\text{norm}} = \mathbf{I} - \mathbf{D}^{-1/2}\mathbf{A}\mathbf{D}^{-1/2}$ and obtain its complete eigendecomposition $\mathbf{L}_{\text{norm}} = \mathbf{U}\mathbf{\Lambda}\mathbf{U}^\top$, where $\mathbf{U} = [\mathbf{u}_1, \mathbf{u}_2, \ldots, \mathbf{u}_{N}]$ are the eigenvectors ordered by increasing eigenvalues $0 = \lambda_1 \leq \lambda_2 \leq \cdots \leq \lambda_{N} \leq 2$.
The eigenvectors corresponding to smaller eigenvalues represent smooth, slowly-varying patterns on the graph (low frequencies), while those associated with larger eigenvalues capture rapidly oscillating patterns (high frequencies).

The bandwidth parameter $R$ is randomly sampled from the uniform distribution $R \sim \text{Uniform}\{3, \lfloor 0.2N \rfloor\}$, ensuring signals are predominantly low-frequency while maintaining sufficient diversity across samples.
We then extract the first $R$ eigenvectors to form the matrix $\mathbf{U}_R = [\mathbf{u}_1, \mathbf{u}_2, \ldots, \mathbf{u}_{R}] \in \mathbb{R}^{N \times R}$.
The spectral coefficient vector $\mathbf{a} \in \mathbb{R}^R$ is sampled independently from a Gaussian distribution $\mathbf{a} \sim \mathcal{N}(\mu \mathbf{1}, \sigma^2 \mathbf{I})$, where the mean is fixed at $\mu = 0$ and the standard deviation $\sigma$ is uniformly sampled from the range $[1.0, 2.0]$.
This generates the clean band-limited signal as $\mathbf{x}^{\text{clean}} = \mathbf{U}_R \mathbf{a}$, which by construction lies in the span of the first $R$ eigenvectors and thus exhibits smooth variation aligned with the graph structure.

To introduce realistic perturbations, we add Gaussian noise $\mathbf{n} \sim \mathcal{N}(\mathbf{0}, \sigma_n^2 \mathbf{I})$ to obtain the observed signal $\mathbf{x} = \mathbf{x}^{\text{clean}} + \mathbf{n}$.
The noise standard deviation is computed as $\sigma_n = \alpha \cdot \sigma$, where the noise level parameter $\alpha$ is uniformly sampled from $[0.05, 0.2]$.
This ensures that the noise magnitude scales proportionally with the signal's coefficient variance, maintaining interpretable signal-to-noise ratios across different samples.

The resulting signal $\mathbf{x}$ has most of its energy in the low-frequency subspace spanned by $\mathbf{U}_R$, with additional high-frequency components contributed by the noise, making it approximately band-limited.

\section{Hyperparameter optimization for graph signal sampling baselines.}
\label{gss-optim}

To ensure a fair comparison between GSS methods and our learning-based approach, we conduct comprehensive hyperparameter optimization for each GSS method on the validation split of the ABL dataset. Since GSS methods are non-parametric and operate on individual graph signals without training, we optimize their hyperparameters using grid search to ensure their best possible performance.

\textbf{Hyperparameter grids.}
For each GSS method, we define a method-specific hyperparameter grid and evaluate all combinations on the validation set. The hyperparameters are selected to minimize the mean squared error (MSE) between the reconstructed signal and the observed (noisy) signal. The hyperparameter spaces are as follows:

\begin{itemize}
    \item \textbf{ESS}: greedy selection parameter $k \in \{1, 2, 3, 4, 5\}$, block size $\in \{1, 2, 3\}$, and bandwidth mode $\in \{\text{exact}, 0.1, 0.2, 0.5, 1.0, 1.5, 2.0, 2.5, 5.0\}$.

    \item \textbf{BSGDA}: regularization parameter $\mu \in \{10^{-4}, 10^{-3}, 10^{-2}, 0.1, 0.5, 1.0, 2.0\}$, convergence threshold $\epsilon \in \{10^{-6}, 10^{-5}, 10^{-4}\}$, localization hops $p \in \{3, 6, 9, 12, 15, 21\}$, and regularization order $\in \{1, 2, 3\}$.

    \item \textbf{FastGSSS}: bandwidth mode $\in \{\text{exact}, 0.1, 0.2, 0.5, 1.0, 1.5, 2.0, 2.5, 5.0\}$, heat kernel width $\nu \in \{25, 50, 75, 100, 150\}$, Chebyshev order $\in \{8, 10, 12, 15, 20\}$, and reconstruction order $\in \{1, 2, 3, 4, 5\}$.

    \item \textbf{RSBS}: bandwidth mode $\in \{\text{exact}, 0.75, 1.0, 1.25, 1.5\}$, number of random vectors for eigenvalue estimation $\in \{1, 3, 5\}$, binary search precision $\epsilon \in \{10^{-3}, 10^{-2}, 10^{-1}\}$, Chebyshev order $\in \{20, 30, 40\}$, reconstruction regularization $\mu \in \{10^{-3}, 10^{-2}, 0.1, 0.5\}$, regularization order $\in \{1, 2\}$, and Laplacian type $\in \{\text{combinatorial}, \text{normalized}\}$.
\end{itemize}

\textbf{Bandwidth mode: Oracle vs. Data-Driven}
A critical hyperparameter across GSS methods is the \textbf{bandwidth mode}, which controls how much oracle information about the true signal bandwidth $R$ is provided to the method:
\begin{itemize}
    \item \textbf{Oracle setting (bandwidth = `exact')}: The method receives the true bandwidth $R$ directly. This represents an optimistic baseline and is the setting we use in our main comparisons to favor GSS methods.

    \item \textbf{Data-driven setting (bandwidth = $\alpha \in \mathbb{R}_+$)}: The method receives an estimated bandwidth $k = \max(2, \lfloor \alpha \cdot M \rfloor)$ where $M$ is the sampling budget. This represents a more realistic scenario where bandwidth must be inferred from problem constraints.
\end{itemize}
In our main experiments, we report results using the oracle `exact' bandwidth mode to provide GSS methods with maximal advantage.

\section{Supplementary Quantitative Metrics.}
\label{appendix:supp_metrics}

Tables~\ref{tab:mse_results} and~\ref{tab:rel_l2_results} report MSE and
relative $\ell_2$ error for all neural models. Classical GSS baselines are
omitted from these tables; their RMSE comparison is the primary metric reported
in Table~\ref{tab:pde_results} of the main text.

\textbf{MSE results (Table~\ref{tab:mse_results}).}
MSE is related to RMSE by $\mathrm{MSE}^{(j)} =
\|\hat{\mathbf{x}}^{(j)}-\mathbf{x}^{(j)}\|_2^2/N$, so the dataset ordering
is identical to RMSE. On Viscoelastic Instability, GWCS achieves
$3.80{\times}10^{-6}$, more than $4{\times}$ lower than the next-best method
(SGAE-50K, $1.88{\times}10^{-5}$), directly quantifying the squared advantage
of the wavelet prior on spectrally structured data. On Turbulent Radiative Layer,
SGAE-99K achieves the best MSE ($27.90$), but its standard deviation
(${\pm}29.39$) exceeds its mean, revealing severe per-sample variation: SGAE-99K
occasionally fails catastrophically on individual turbulent snapshots while still
achieving the best aggregate error. GWCS shows much lower variance
($51.26{\pm}15.49$), making it more predictable across samples. On Kolmogorov
Flow, SGAE-50K is best ($0.0861$) with GWCS ($0.1019$) close behind; on Dynamic
Stall, SGAE-99K dominates ($6.70$) while GWCS ($85.82$) is at GXN level
($90.88$).

\textbf{Relative $\ell_2$ error (Table~\ref{tab:rel_l2_results}).}
The relative $\ell_2$ error $\mathrm{relL2}^{(j)} = \|\hat{\mathbf{x}}^{(j)}-\mathbf{x}^{(j)}\|_2/\|\mathbf{x}^{(j)}\|_2$ is scale-invariant; values below 1.0 indicate error smaller than the signal
norm.

On Viscoelastic Instability, GWCS achieves the best relative $\ell_2$ ($0.390$), confirming that the wavelet prior captures the largest fraction of signal energy. The SGAE variants show high relative $\ell_2$ ($0.75$--$0.92$) with very large standard deviations (SGAE-50K: ${\pm}0.94$); this is a numerical artifact, some Viscoelastic test samples have near-zero signal norm, inflating the denominator. MSE (Table~\ref{tab:mse_results}) is the more reliable metric for this dataset.

On Turbulent Radiative Layer and Kolmogorov Flow, the ordering matches RMSE exactly: SGAE-99K leads Turbulent ($0.085$) and SGAE-50K leads Kolmogorov ($0.306$), with GWCS close behind in both cases.

On Dynamic Stall, all relative $\ell_2$ values appear small in absolute terms ($0.025$--$0.095$) because the pressure signal has a large non-zero mean (${\approx}98{,}160$\,Pa) that dominates the norm, making the denominator large. The relative ordering is nonetheless preserved (SGAE-99K best at $0.025$, GWCS
and GXN both near $0.092$--$0.095$), consistent with the RMSE ranking.

\begin{table}[h]
\centering
\caption{MSE (mean\,$\pm$\,std, physical units$^2$) on four PDE datasets at compression ratio $c=5\%$. \textbf{Bold}: best; \underline{underline}: second-best.}
\label{tab:mse_results}
\small
\setlength{\tabcolsep}{3pt}
\begin{tabular}{lcccc}
\toprule
\textbf{Method} &
  \textbf{Viscoelastic} &
  \textbf{Turbulent} &
  \textbf{Kolmogorov} &
  \textbf{Dynamic Stall} \\
\midrule
GXN
  & $4.21{\times}10^{-5}$$\pm$$5.02{\times}10^{-5}$
  & $1724.1\pm124.8$
  & $0.737\pm0.190$
  & $90.88\pm34.92$ \\
SGAE-50K
  & \underline{$1.88{\times}10^{-5}$$\pm$$2.53{\times}10^{-5}$}
  & \underline{$40.06\pm43.60$}
  & $\mathbf{0.0861\pm0.0377}$
  & \underline{$10.98\pm5.97$} \\
SGAE-99K
  & $2.28{\times}10^{-5}$$\pm$$3.17{\times}10^{-5}$
  & $\mathbf{27.90\pm29.39}$
  & $0.110\pm0.038$
  & $\mathbf{6.70\pm4.23}$ \\
\midrule
\textbf{GWCS}
  & $\mathbf{3.80{\times}10^{-6}\pm4.46{\times}10^{-6}}$
  & $51.26\pm15.49$
  & \underline{$0.102\pm0.026$}
  & $85.82\pm30.91$ \\
\bottomrule
\end{tabular}
\end{table}

\begin{table}[h]
\centering
\caption{Relative $\ell_2$ error (mean\,$\pm$\,std) on four PDE datasets at compression ratio $c=5\%$. \textbf{Bold}: best; \underline{underline}: second-best.}
\label{tab:rel_l2_results}
\small
\setlength{\tabcolsep}{3pt}
\begin{tabular}{lcccc}
\toprule
\textbf{Method} &
  \textbf{Viscoelastic} &
  \textbf{Turbulent} &
  \textbf{Kolmogorov} &
  \textbf{Dynamic Stall} \\
\midrule
GXN
  & $0.947\pm0.039$
  & $0.737\pm0.018$
  & $0.910\pm0.018$
  & $0.095\pm0.020$ \\
SGAE-50K
  & $0.923\pm0.936$
  & \underline{$0.102\pm0.050$}
  & $\mathbf{0.306\pm0.032}$
  & \underline{$0.033\pm0.009$} \\
SGAE-99K
  & \underline{$0.753\pm0.336$}
  & $\mathbf{0.085\pm0.042}$
  & $0.357\pm0.081$
  & $\mathbf{0.025\pm0.008}$ \\
\midrule
\textbf{GWCS}
  & $\mathbf{0.390\pm0.240}$
  & $0.126\pm0.021$
  & \underline{$0.339\pm0.013$}
  & $0.092\pm0.018$ \\
\bottomrule
\end{tabular}
\end{table}

\section{Qualitative Reconstruction.}
\label{appendix:qual_all}
\begin{figure*}[!ht]
  \centering
  \includegraphics[width=0.95\textwidth]{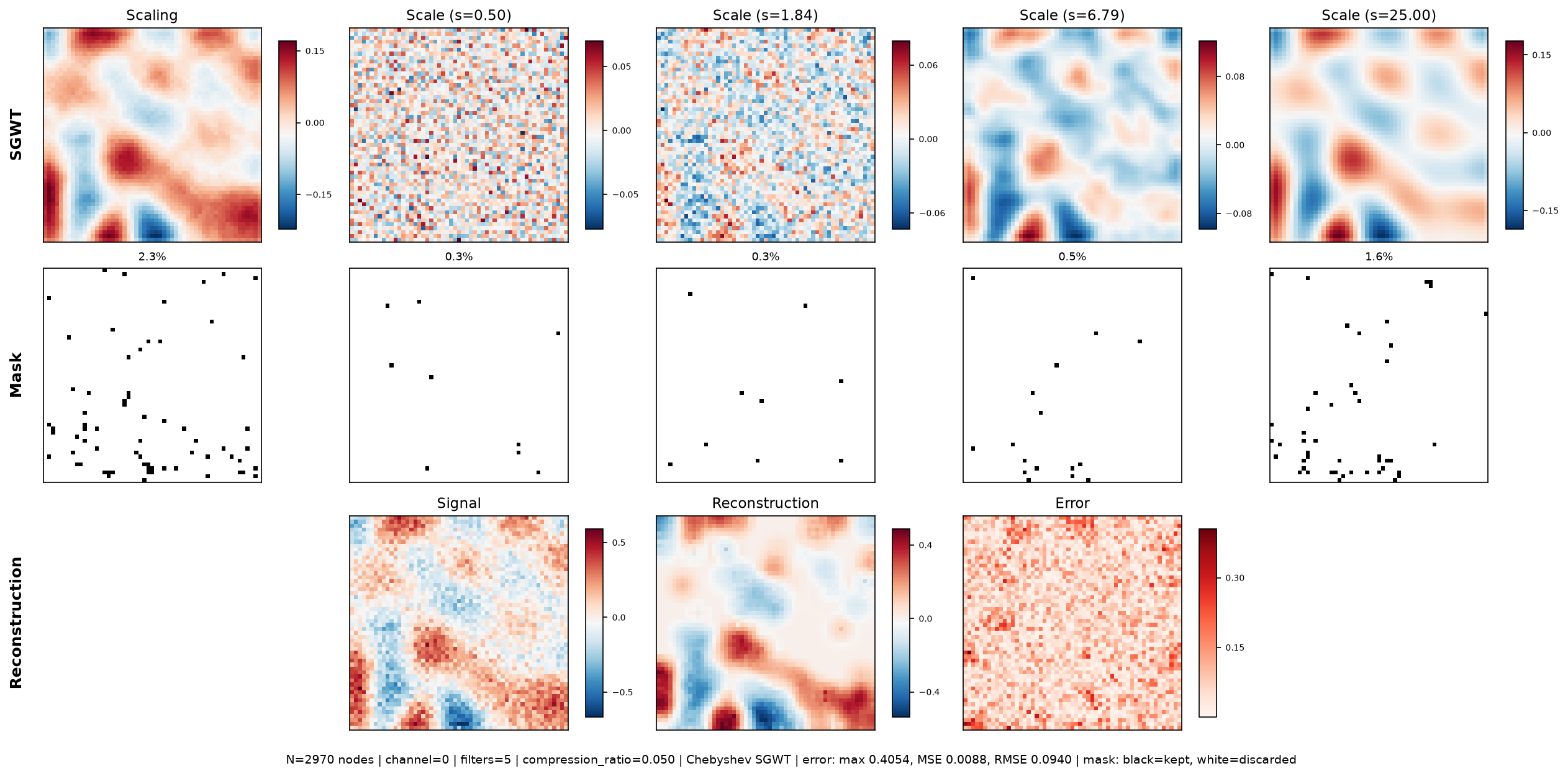}
  \caption{GWCS pipeline visualization on a representative grid-graph ABL signal (test sample~407, $N=2970$ nodes, RMSE\,$=\,0.0940$): SGWT decomposition (top), MLIS sparsification mask per scale (middle), and NIGWT reconstruction with pointwise error (bottom).}
  \label{fig:app_grid_1}
\end{figure*}

\begin{figure*}[!ht]
  \centering
  \includegraphics[width=0.95\textwidth]{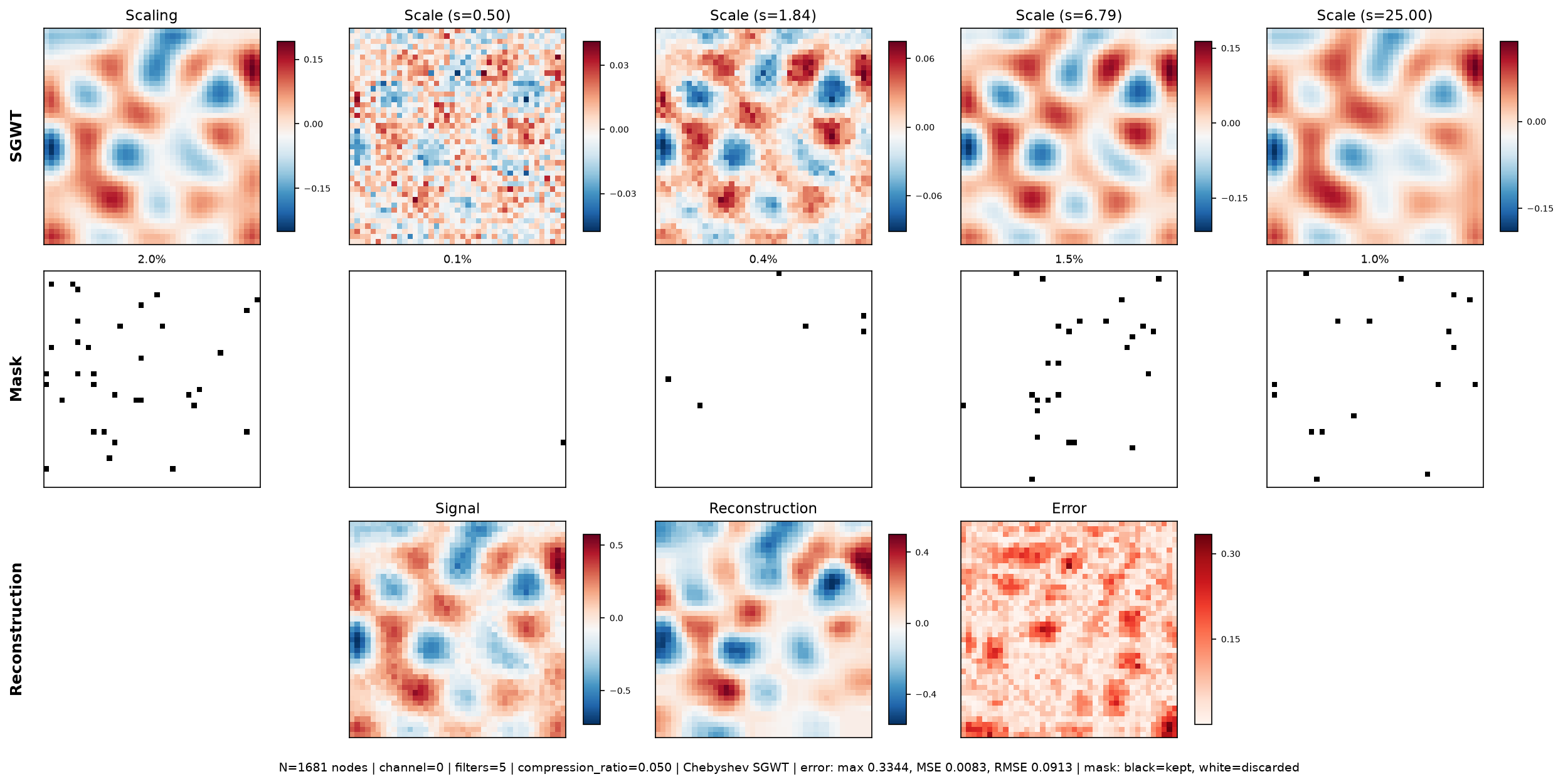}
  \caption{GWCS pipeline visualization on a representative grid-graph ABL signal (test sample~461, $N=1681$ nodes, RMSE\,$=\,0.0913$): SGWT decomposition (top), MLIS sparsification mask per scale (middle), and NIGWT reconstruction with pointwise error (bottom).}
  \label{fig:app_grid_2}
\end{figure*}

\begin{figure*}[!ht]
  \centering
  \includegraphics[width=0.95\textwidth]{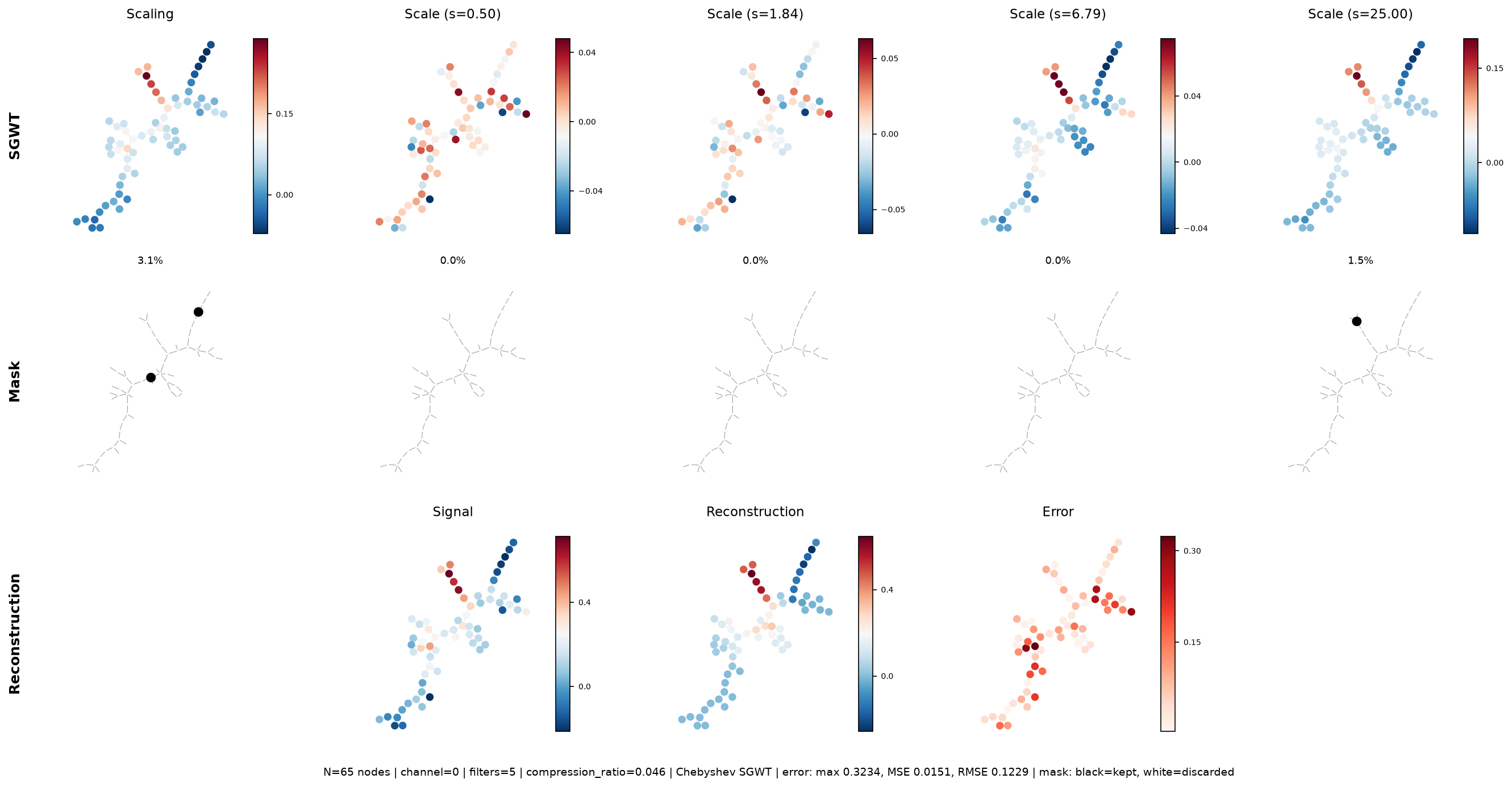}
  \caption{GWCS pipeline visualization on a representative Erd\H{o}s--R\'{e}nyi-graph ABL signal (test sample~107, $N=65$ nodes, RMSE\,$=\,0.1229$): SGWT decomposition (top), MLIS sparsification mask per scale (middle), and NIGWT reconstruction with pointwise error (bottom).}
  \label{fig:app_er_1}
\end{figure*}

\begin{figure*}[!ht]
  \centering
  \includegraphics[width=0.95\textwidth]{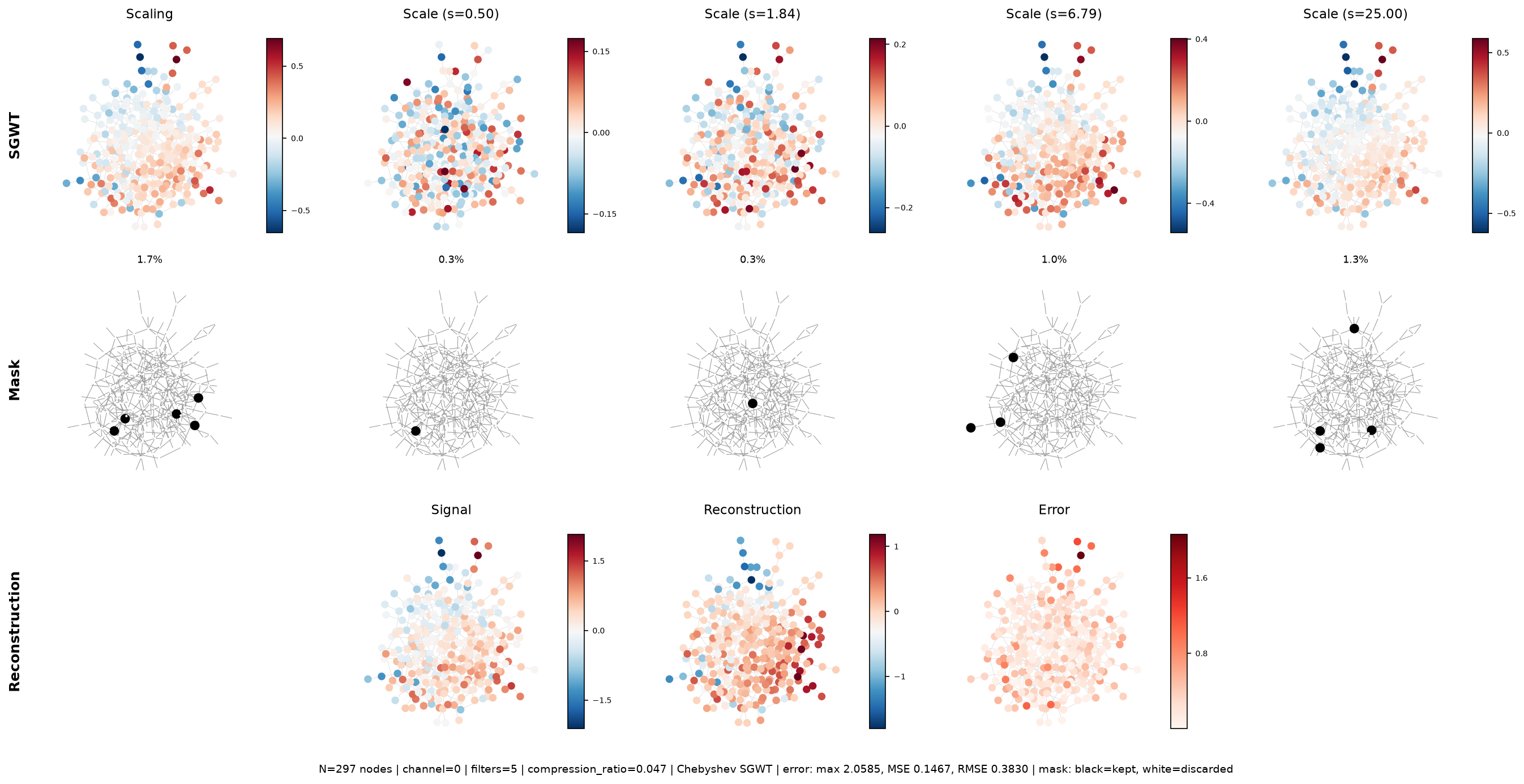}
  \caption{GWCS pipeline visualization on a representative Erd\H{o}s--R\'{e}nyi-graph ABL signal (test sample~261, $N=297$ nodes, RMSE\,$=\,0.3830$): SGWT decomposition (top), MLIS sparsification mask per scale (middle), and NIGWT reconstruction with pointwise error (bottom).}
  \label{fig:app_er_3}
\end{figure*}

\begin{figure*}[!ht]
  \centering
  \subfigure[GWCS (Ours) --- RMSE\,$=\,0.319$]{%
    \includegraphics[width=0.7\textwidth]{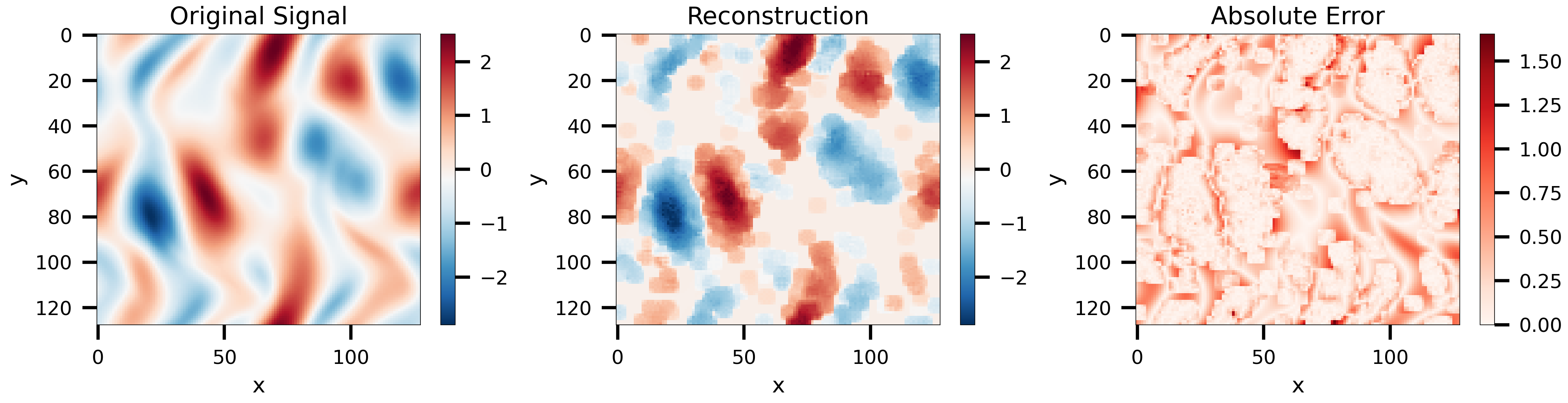}}\\[-4pt]
  \subfigure[SGAE-99K --- RMSE\,$=\,0.328$]{%
    \includegraphics[width=0.7\textwidth]{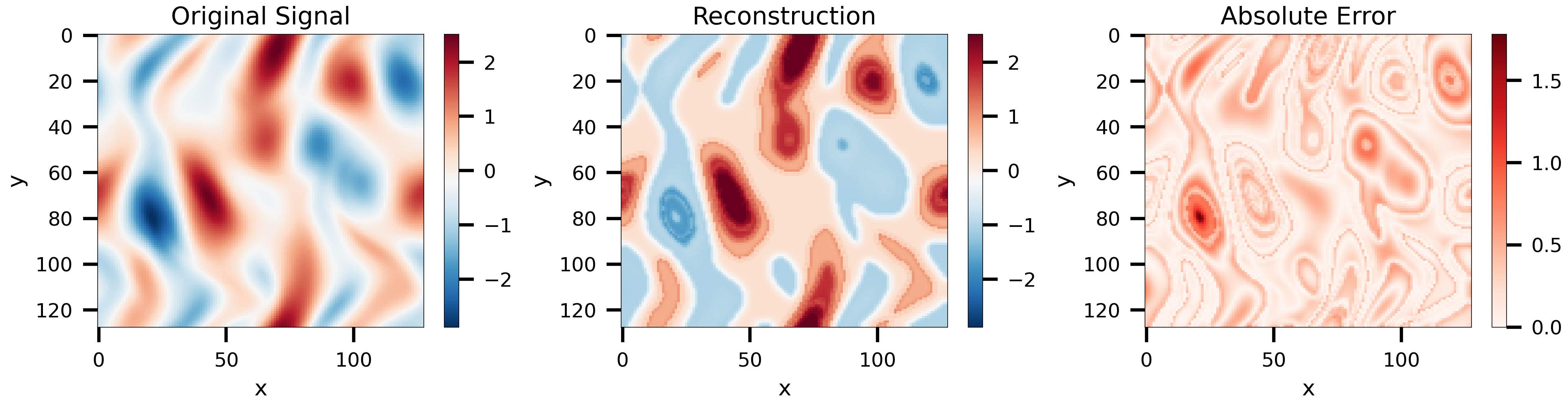}}\\[-4pt]
  \subfigure[GXN --- RMSE\,$=\,0.851$]{%
    \includegraphics[width=0.7\textwidth]{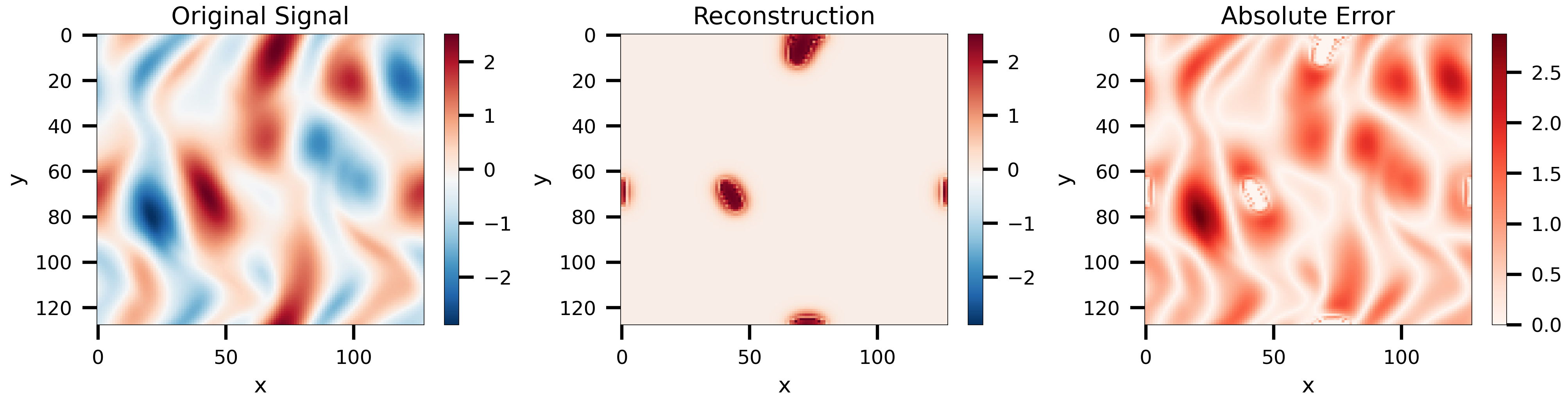}}\\[-4pt]
  \subfigure[FastGSSS --- RMSE\,$=\,0.903$]{%
    \includegraphics[width=0.7\textwidth]{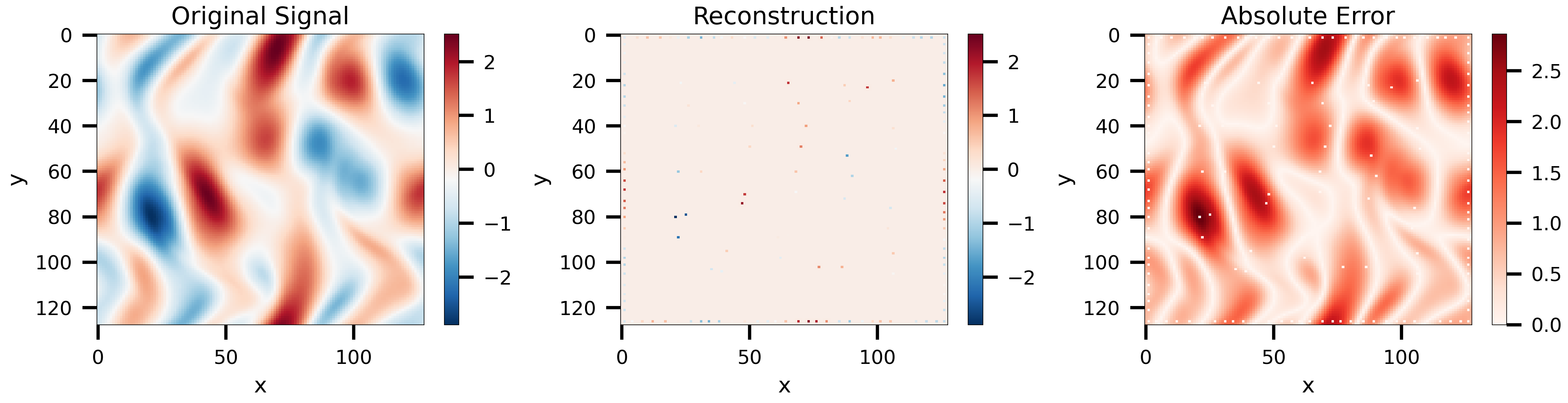}}\\[-4pt]
  \subfigure[BSGDA --- RMSE\,$=\,0.545$]{%
    \includegraphics[width=0.7\textwidth]{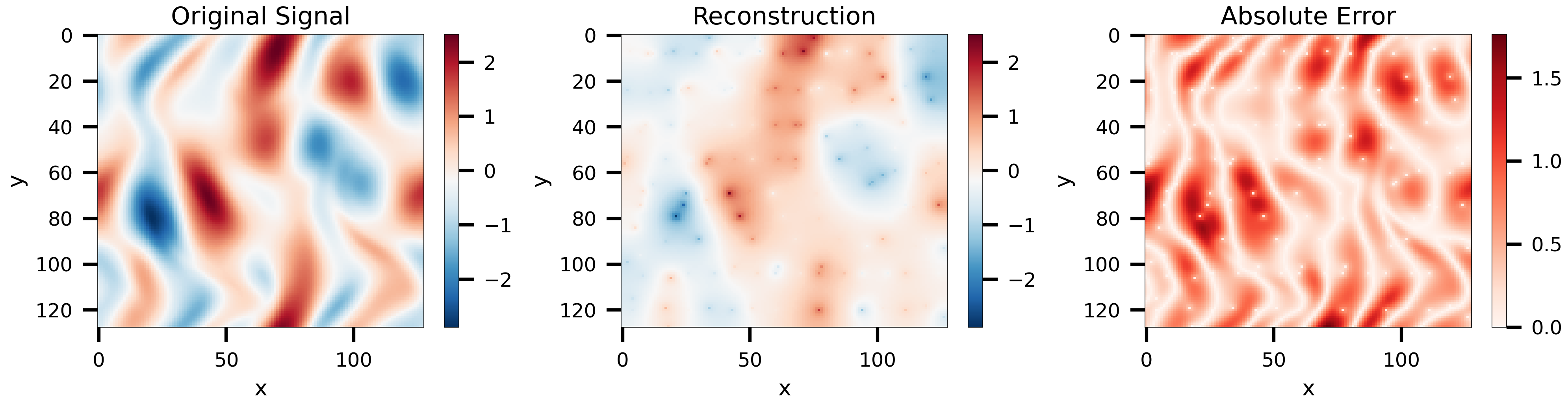}}\\[-4pt]
  \subfigure[RSBS --- RMSE\,$=\,0.609$]{%
    \includegraphics[width=0.7\textwidth]{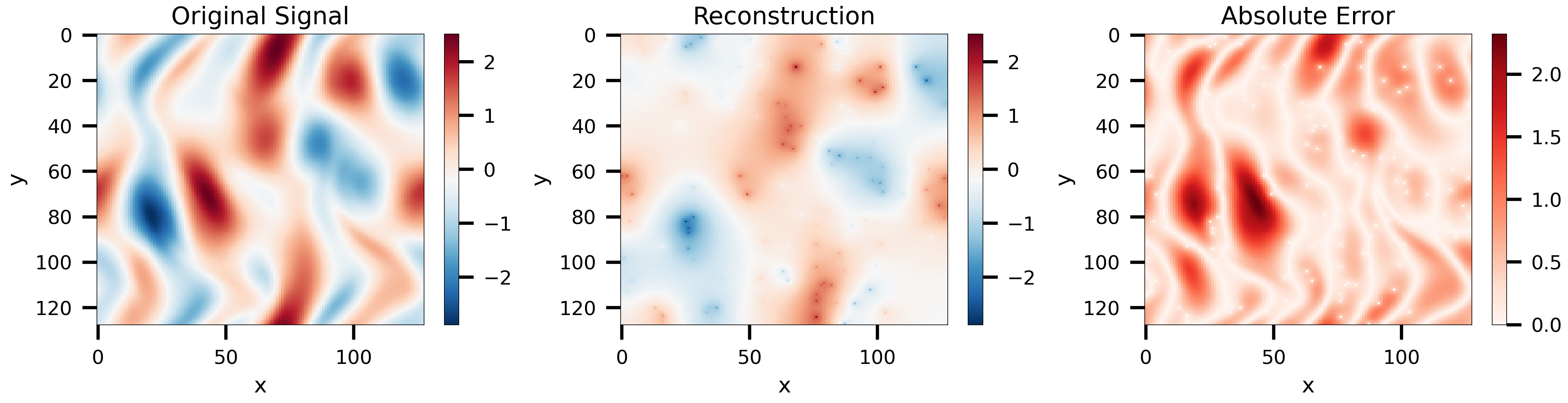}}
  \caption{Qualitative reconstruction of the Kolmogorov Flow velocity$_x$ field
  (test sample~47).}
  \label{fig:qual_kolmogorov}
\end{figure*}

\begin{figure*}[!ht]
  \centering
  \subfigure[GWCS (Ours) --- RMSE\,$=\,9119$]{%
    \includegraphics[width=0.7\textwidth]{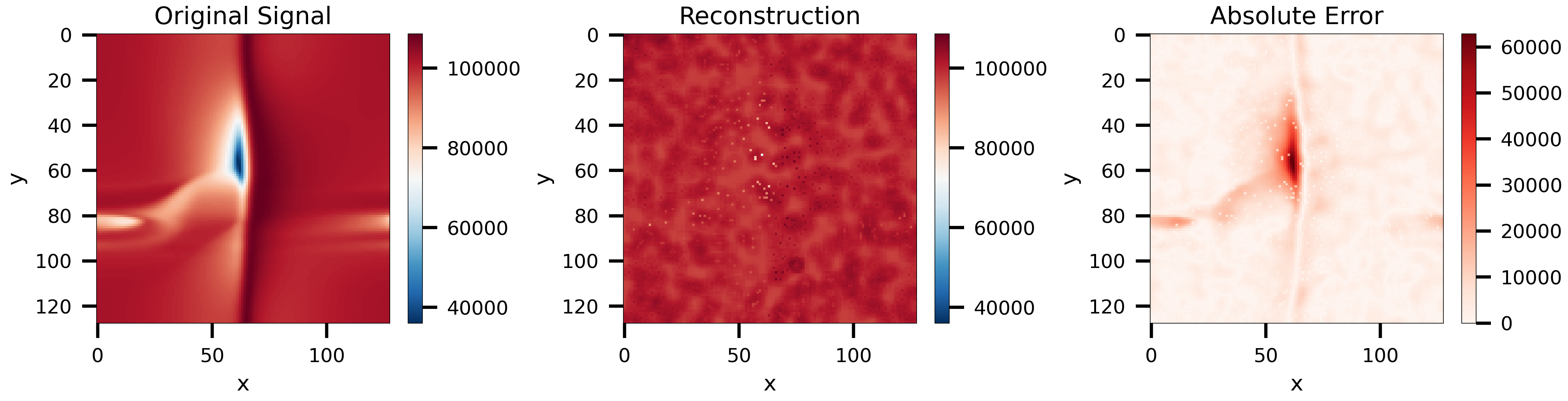}}\\[-4pt]
  \subfigure[SGAE-99K --- RMSE\,$=\,2480$]{%
    \includegraphics[width=0.7\textwidth]{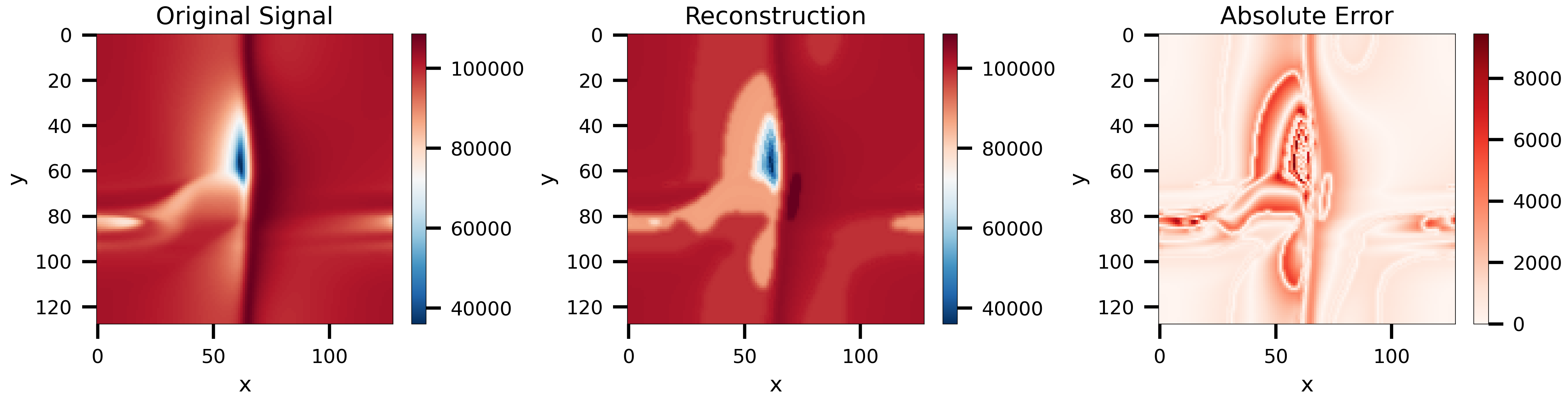}}\\[-4pt]
  \subfigure[GXN --- RMSE\,$=\,9350$]{%
    \includegraphics[width=0.7\textwidth]{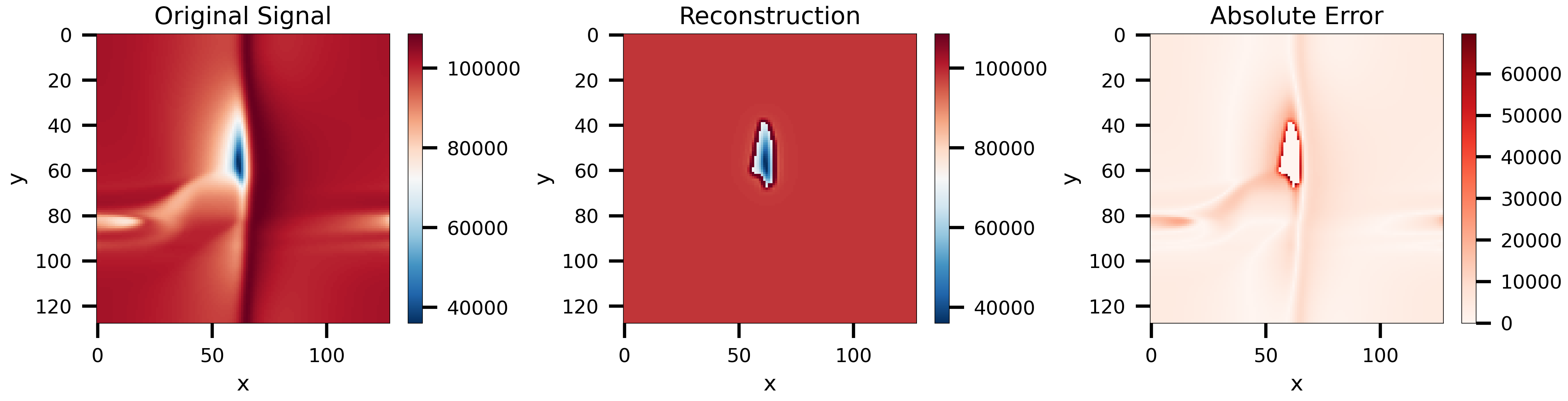}}\\[-4pt]
  \subfigure[FastGSSS --- RMSE\,$=\,6788$ (sample 48)]{%
    \includegraphics[width=0.7\textwidth]{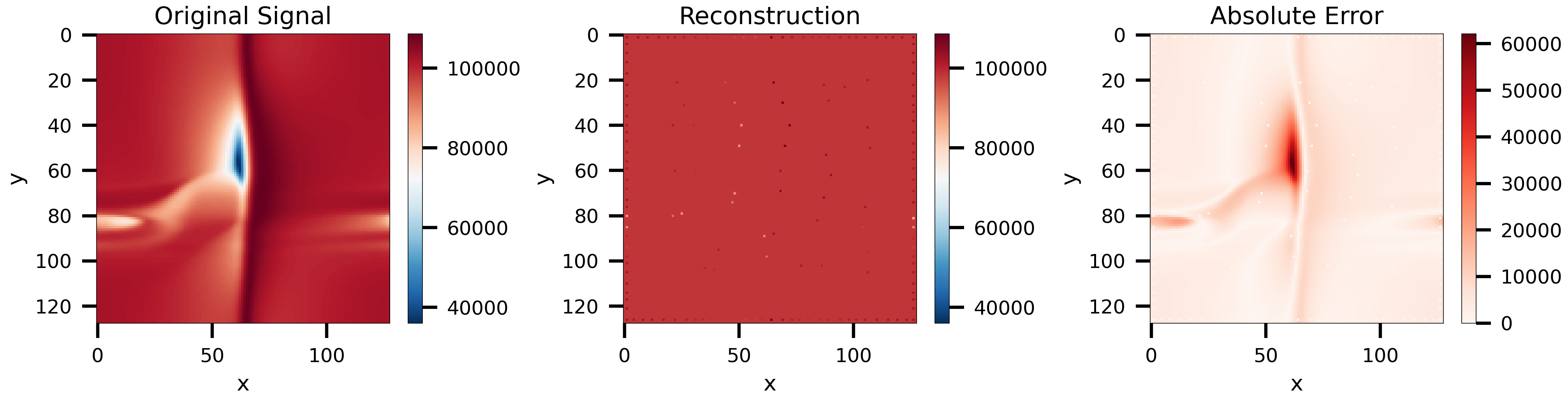}}\\[-4pt]
  \subfigure[BSGDA --- RMSE\,$=\,3954$ (sample 48)]{%
    \includegraphics[width=0.7\textwidth]{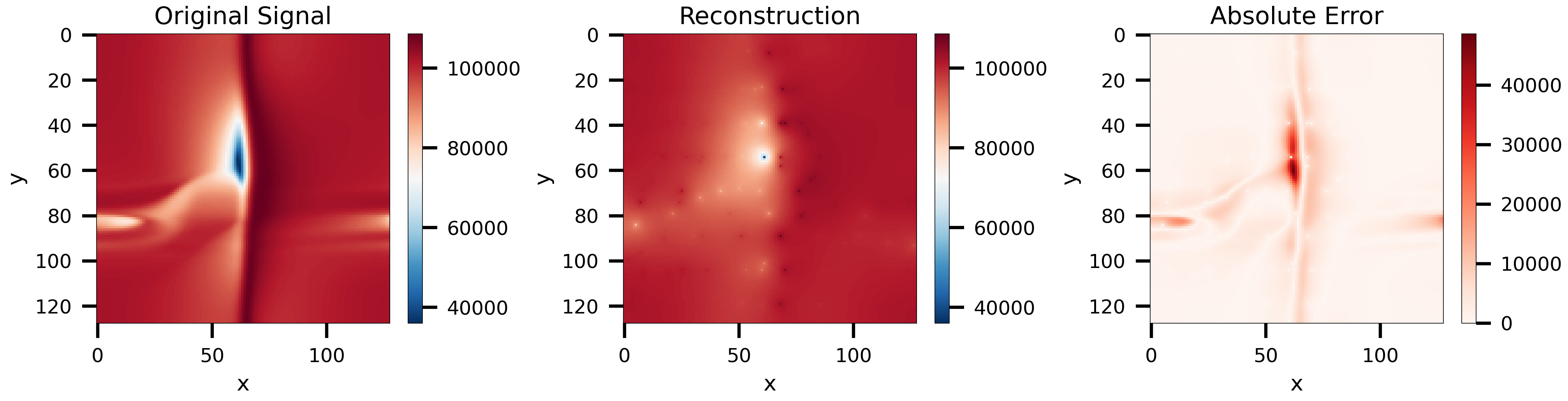}}\\[-4pt]
  \subfigure[RSBS --- RMSE\,$=\,5033$ (sample 48)]{%
    \includegraphics[width=0.7\textwidth]{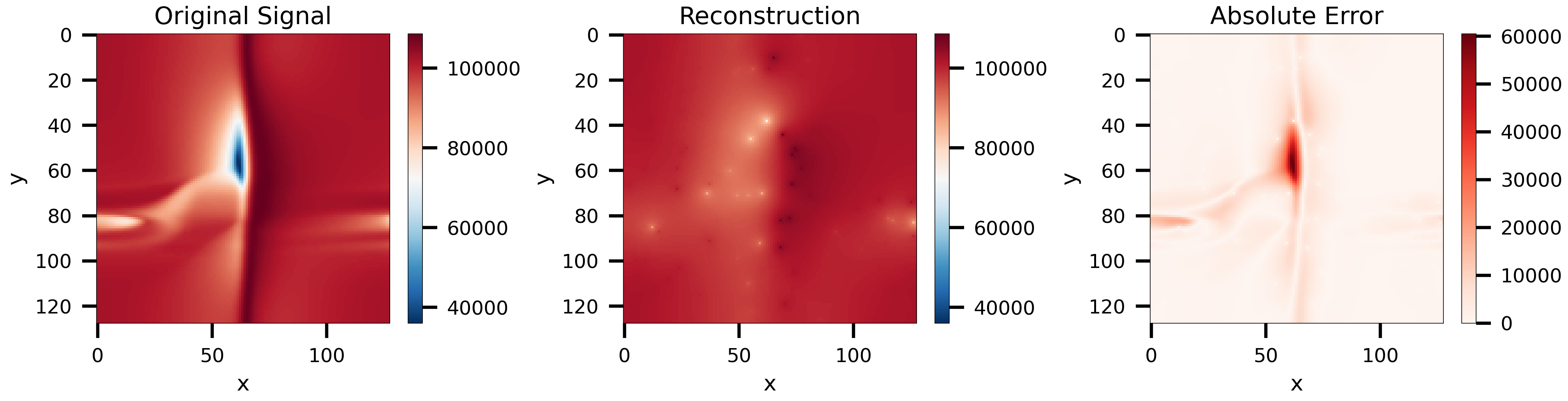}}
  \caption{Qualitative reconstruction of the Dynamic Stall pressure field
  (test sample~47).}
  \label{fig:qual_dynamic_stall}
\end{figure*}

\begin{figure*}[!ht]
  \centering
  \subfigure[GWCS (Ours) --- RMSE\,$=\,0.0016$]{%
    \includegraphics[width=0.7\textwidth]{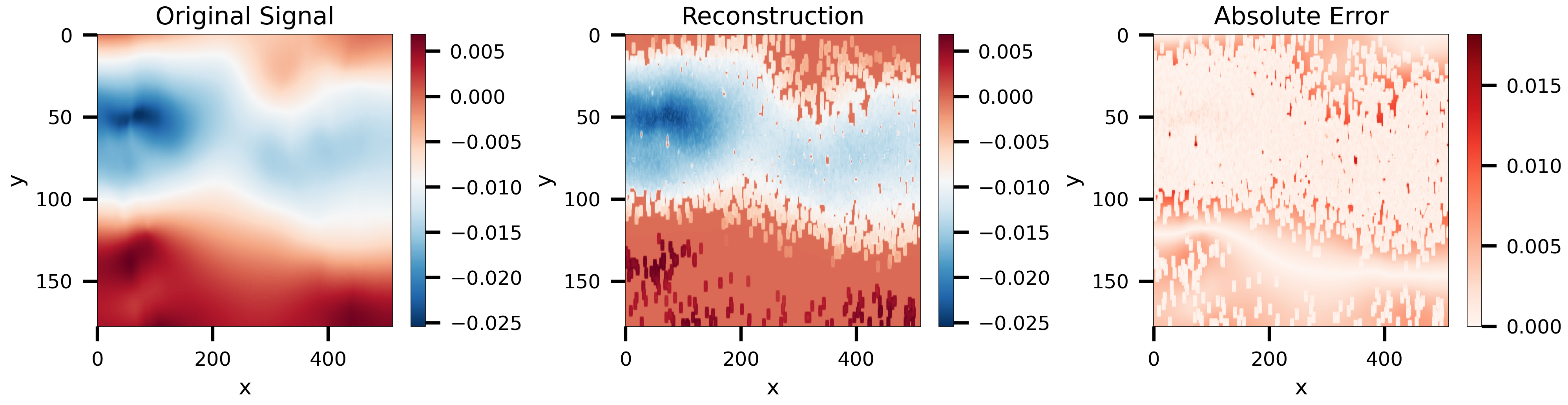}}\\[-4pt]
  \subfigure[SGAE-99K --- RMSE\,$=\,0.0037$]{%
    \includegraphics[width=0.7\textwidth]{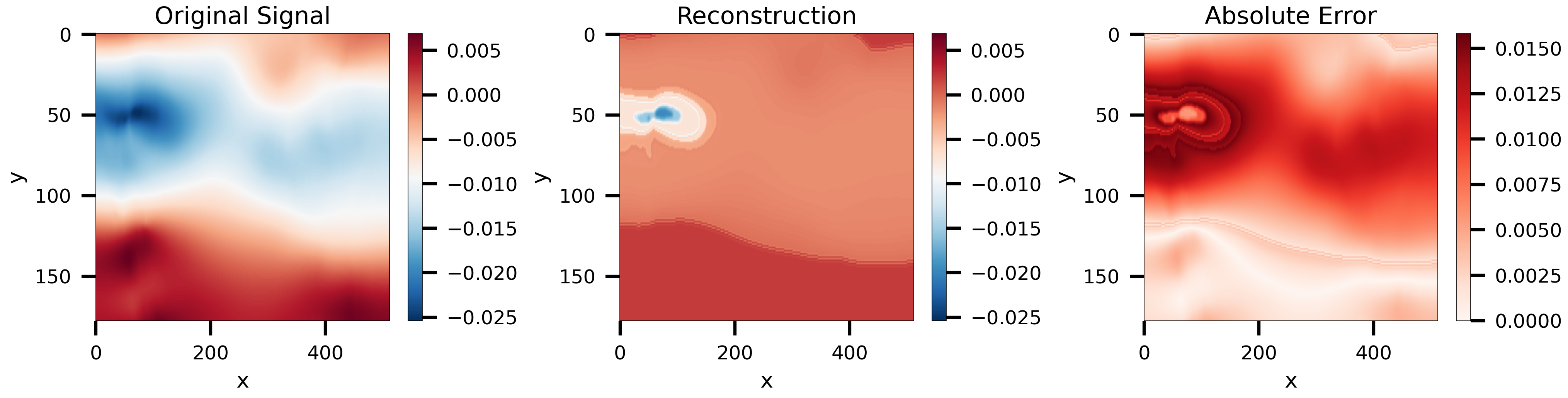}}\\[-4pt]
  \subfigure[GXN --- RMSE\,$=\,0.0052$]{%
    \includegraphics[width=0.7\textwidth]{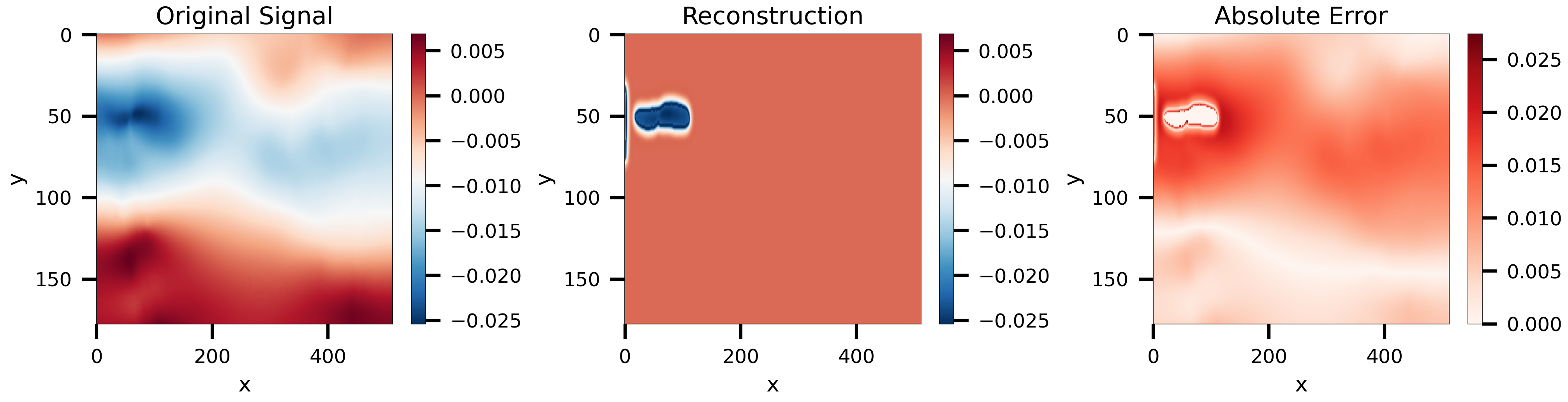}}
  \caption{Qualitative reconstruction of the Viscoelastic Instability pressure
  field (test sample~47).}
  \label{fig:qual_viscoelastic}
\end{figure*}

Figures~\ref{fig:app_grid_1}--\ref{fig:app_er_3} visualize the SGWT coefficients, MLIS sparsification mask, and NIGWT reconstruction for four representative ABL test samples over grid and Erd\H{o}s--R\'{e}nyi graphs at target compression ratio $c=5\%$. 
Figures~\ref{fig:qual_kolmogorov}--\ref{fig:qual_viscoelastic} show qualitative
reconstructions of single test sample for Turbulent Radiative Layer, Kolmogorov Flow, Dynamic Stall, and Viscoelastic Instability, respectively. Each row shows (left to right) the ground-truth signal, the reconstructed signal, and the pointwise absolute error. Neural methods (GWCS, SGAE-99K, GXN) are shown for all four datasets; classical baselines (FastGSSS, BSGDA, RSBS) are shown for Turbulent, Kolmogorov, and Dynamic Stall only, as classical evaluation of the Viscoelastic dataset was not performed due to prohibitive per-sample computational cost in terms of time and memory.

\begin{figure*}[!ht]
  \centering
  \subfigure[GWCS (Ours) --- RMSE\,$=\,6.90$]{%
    \includegraphics[width=0.7\textwidth]{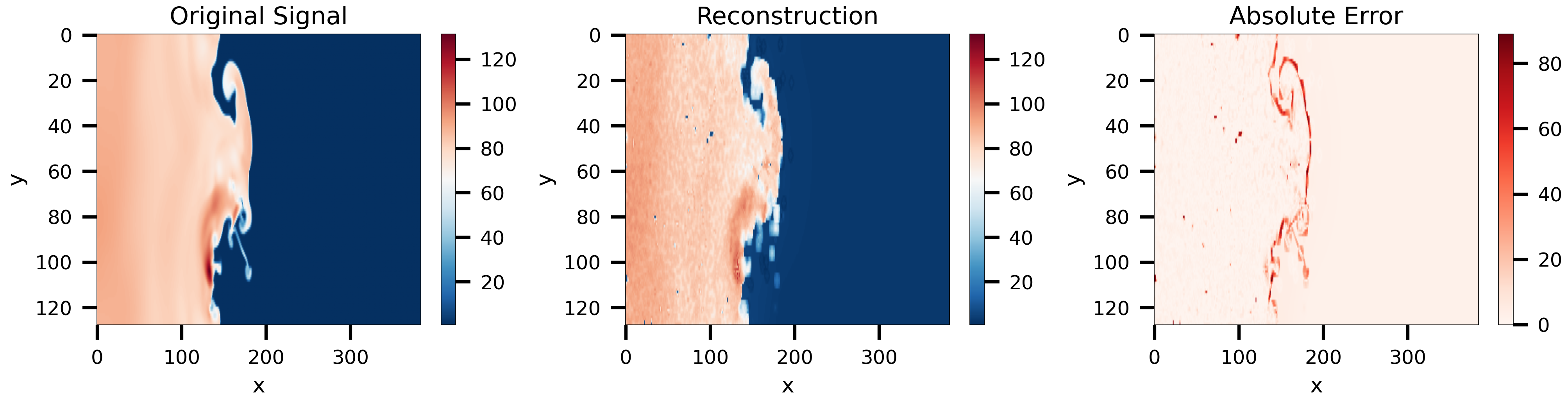}}\\[-4pt]
  \subfigure[SGAE-99K --- RMSE\,$=\,9.32$]{%
    \includegraphics[width=0.7\textwidth]{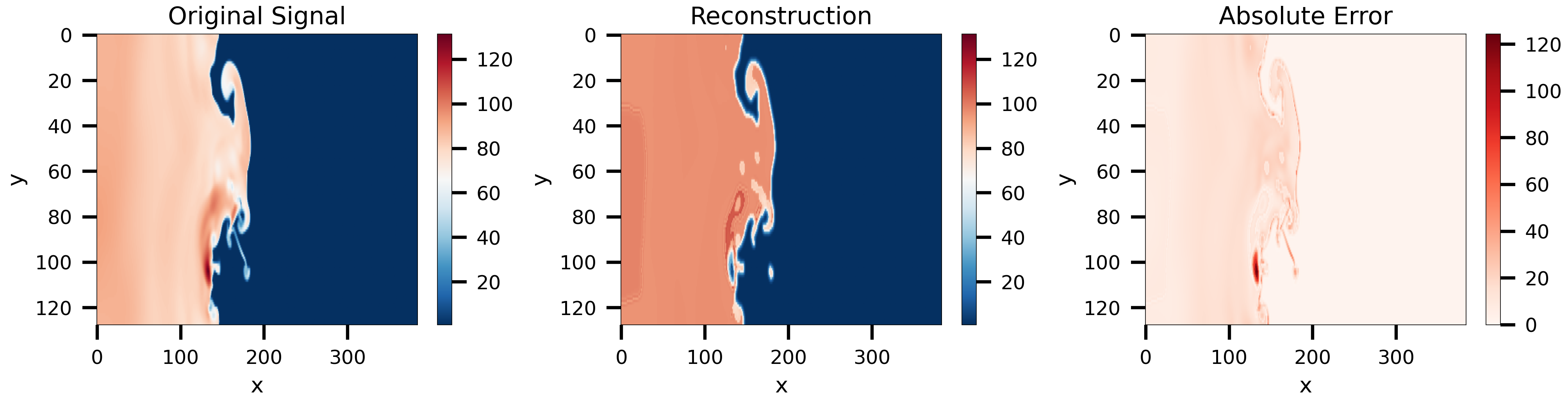}}\\[-4pt]
  \subfigure[GXN --- RMSE\,$=\,38.55$]{%
    \includegraphics[width=0.7\textwidth]{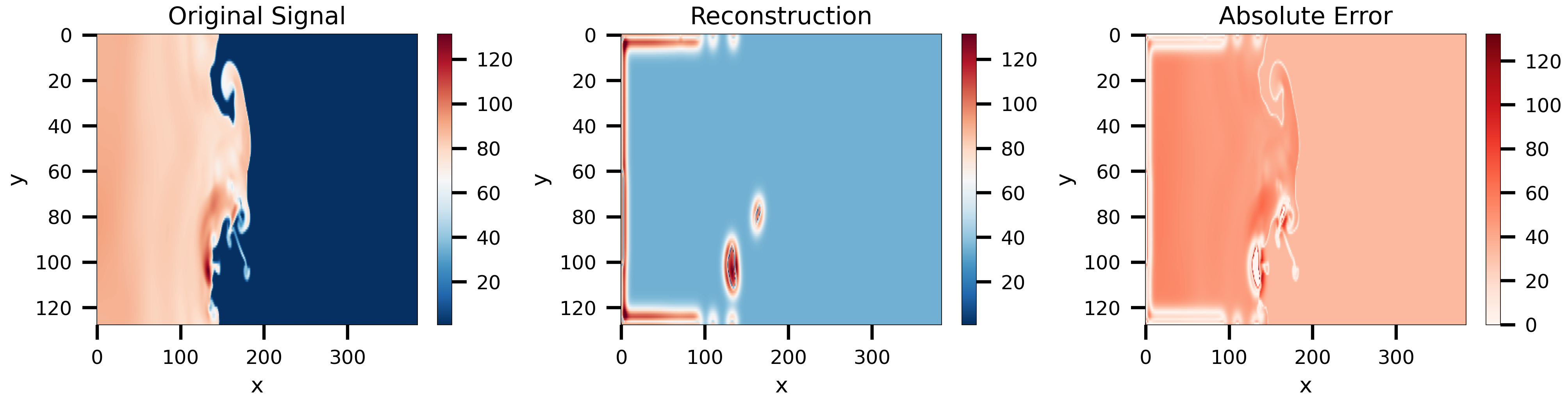}}\\[-4pt]
  \subfigure[FastGSSS --- RMSE\,$=\,40.47$ (sample 48)]{%
    \includegraphics[width=0.7\textwidth]{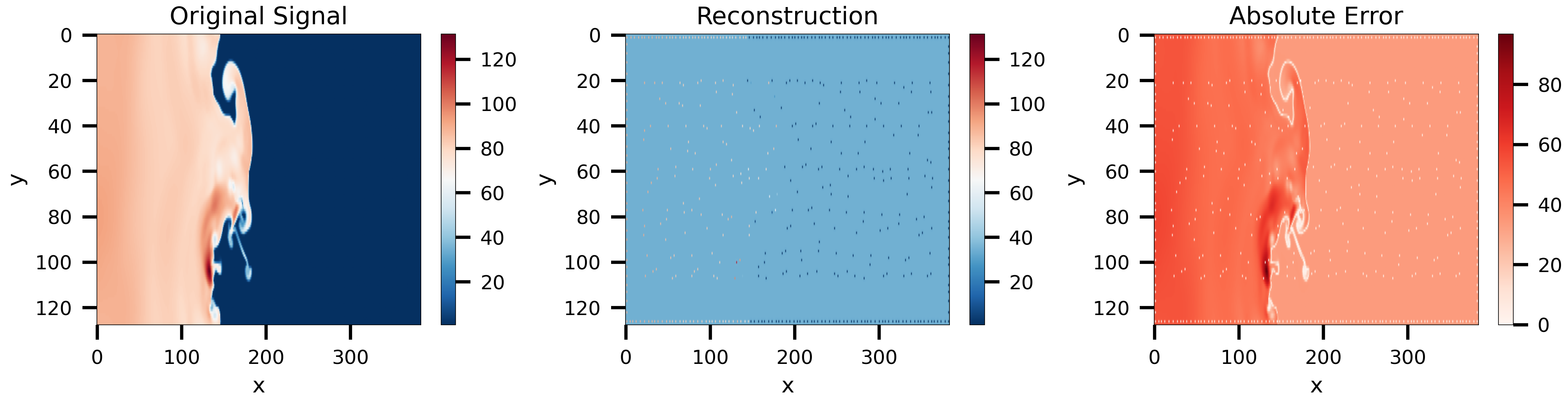}}\\[-4pt]
  \subfigure[BSGDA --- RMSE\,$=\,8.88$ (sample 48)]{%
    \includegraphics[width=0.7\textwidth]{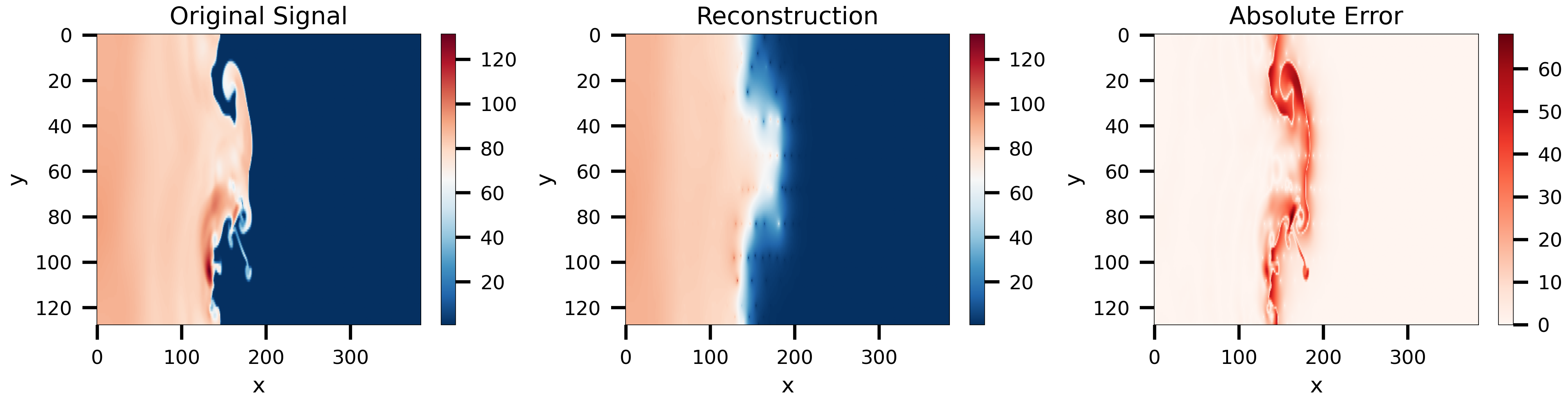}}\\[-4pt]
  \subfigure[RSBS --- RMSE\,$=\,9.25$ (sample 48)]{%
    \includegraphics[width=0.7\textwidth]{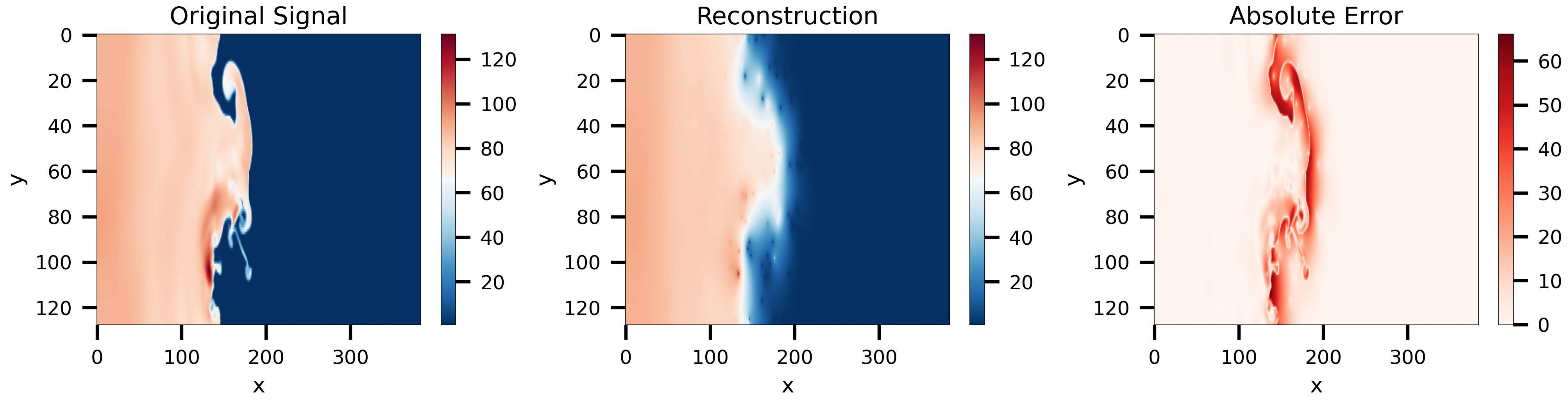}}
  \caption{Qualitative reconstruction of the Turbulent Radiative Layer density
  field (test sample~47).}
  \label{fig:qual_turbulent}
\end{figure*}

\section{Code Availability}
\label{app:code_avail}
The code is available at the GitHub repository: \url{https://github.com/amrhssn/gwcs/}.
\end{document}